\documentclass[twoside,11pt]{article}

\pdfoutput=1

\usepackage{arxiv}

\usepackage{hyperref}

\usepackage[round]{natbib}

\usepackage[utf8]{inputenc} 

\usepackage{amsmath} 

\usepackage{enumerate} 

\usepackage{siunitx} 

\usepackage{multirow} 

\newcommand{\nlabel}[1]{\addtocounter{equation}{1}\tag{\theequation}\label{#1}} 

\begin{document}

\title{Sparse-Group Bayesian Feature Selection Using Expectation Propagation for Signal Recovery and Network Reconstruction}

\author{\name Edgar Steiger \email steige\_e@molgen.mpg.de \\
       \name Martin Vingron \email vingron@molgen.mpg.de \\
       \addr Max Planck Institute for Molecular Genetics\\
       Ihnestr. 63-73, D-14195 Berlin, Germany}

\maketitle

\begin{abstract}
We present a Bayesian method for feature selection in the presence of grouping information with sparsity on the between- and within group level. Instead of using a stochastic algorithm for parameter inference, we employ expectation propagation, which is a deterministic and fast algorithm.

Available methods for feature selection in the presence of grouping information have a number of short-comings: on one hand, lasso methods, while being fast, underestimate the regression coefficients and do not make good use of the grouping information, and on the other hand, Bayesian approaches, while accurate in parameter estimation, often rely on the stochastic and slow Gibbs sampling procedure to recover the parameters, rendering them infeasible e.g.\ for gene network reconstruction. Our approach of a Bayesian sparse-group framework with expectation propagation enables us to not only recover accurate parameter estimates in signal recovery problems, but also makes it possible to apply this Bayesian framework to large-scale network reconstruction problems.

The presented method is generic but in terms of application we focus on gene regulatory networks. We show on simulated and experimental data that the method constitutes a good choice for network reconstruction regarding the number of correctly selected features, prediction on new data and reasonable computing time. 
\end{abstract}


\section{Introduction}

Feature selection problems with grouped features arise in many statistical applications, for example when prior knowledge about the features is available. In the presence of grouping information, feature selection methods should consider two levels of sparsity, that is between-group sparsity and within-group sparsity. Between-group sparsity discards whole groups of features that bear as a group no influence on the dependent variable. Within-group sparsity chooses within one group of features the ones that are explanatory and discards the remaining features. To take this two-level sparsity into account, the standard linear regression model with feature selection \citep[Chapter 13]{murphy_machine_2012} needs to be extended. 

In general, feature selection can be approached from the frequentist or the Bayesian viewpoint. The frequentist approach infers point estimates of the parameters, based on samples, as approximations to the universal true parameters, while the Bayesian approach treats the parameters as stochastic entities themselves, and as such gives probabilistic interpretations of parameters based on prior beliefs in the framework of Bayes' theorem \citep[Section 1.2.3]{bishop_pattern_2006}. In feature selection, the frequentist approach is mostly represented by the lasso method \citep{tibshirani_regression_1996}. The group lasso \citep{yuan_model_2006} takes between-group sparsity into account, while the sparse-group lasso \citep{friedman_note_2010,simon_sparse-group_2013} considers both between-group sparsity and within-group sparsity. 

An example for a Bayesian approach to feature selection is the spike-and-slab method \citep{mitchell_bayesian_1988}. \citet{xu_bayesian_2015} present different Bayesian models for feature selection in the presence of grouped features. In most instances, the application of the Bayesian approach relies on Gibbs sampling, which is an algorithm to infer parameters within the Bayesian framework \citep{geman_stochastic_1984}, but Gibbs sampling needs to run for long times to give reliable results. As such, the Bayesian approach was ill-suited for some tasks like large-scale gene network inference.  An alternative to Gibbs sampling is expectation propagation \citep{minka_expectation_2001}, a deterministic algorithm not relying on stochastic sampling. Expectation propagation can potentially decrease the run-time of a Bayesian approach at the cost of a larger mathematical overhead and a less straight-forward implementation. \citet{hernandez-lobato_generalized_2013} introduced expectation propagation updates for the spike-and-slab model with between-group sparsity, but an extension that takes within-group sparsity into account remained an open problem.

Here we present a Bayesian framework for feature selection when features are grouped and are sparse on the between- and within-group level, along with closed-form solutions of the corresponding expectation propagation parameter updates, and provide an efficient implementation as well. This allows us to plug the method into the framework of neighborhood selection \citep{meinshausen_high-dimensional_2006} for large-scale network reconstruction, which was restricted to the use of lasso methods before. Furthermore we apply and compare our method on real experimental data to reconstruct gene regulatory networks.

\section{Sparse-Group Bayesian Feature Selection with Expectation Propagation}

In this section we present a Bayesian model and algorithm for feature selection that is sparse on the between- and within-group level. Consider a linear regression model
\begin{align*}
\nlabel{eq:linmatrix}
y = X \cdot \beta + \varepsilon,
\end{align*}
with response vector $y \in \mathbb{R}^M$, vector of coefficients $\beta \in \mathbb{R}^N$, noise vector $\varepsilon \in \mathbb{R}^M$ and data matrix $X \in \mathbb{R}^{M\times N}$. $\varepsilon=(\varepsilon_{m})_{m=1}^M$ is a vector of independent, identically distributed variables (errors) distributed according to $\varepsilon_m \sim \mathcal{N}(0,\sigma_0^2)$. Equation \eqref{eq:linmatrix} is the matrix notation of the following equation:
\begin{align*}
\nlabel{eq:linsum}
y_{m} = \sum_{n=1}^{N} \beta_{n} \cdot x_{mn}+\varepsilon_{m} \textnormal{, where } m=1,\ldots,M.
\end{align*}
 If we assume that most $\beta_n$ are effectively equal to zero, we have the scenario of sparse regression or feature selection. Furthermore, we consider that there is an inherent grouping $\mathcal{G}$ of the variables $(X_1, \ldots, X_N)$. A grouping of the features or equivalent, a map $\mathcal{G} : \lbrace 1,\ldots, N \rbrace \rightarrow \lbrace 1, \ldots, G \rbrace$ with $G$ the number of groups, is an additional information about the structure of the data. The map $\mathcal{G}$ assigns to every feature (respectively its index $n$) a group (respectively group index $g$): $\mathcal{G}(n)=g$. If $\mathcal{G}$ is injective, every feature is mapped to its own group and thus the mapping reduces to a non-informative grouping. For a given group $g$ we denote by $\mathcal{G}^{-1}(g)$ the set of indices $\mathcal{G}^{-1}(g) \subseteq \lbrace 1,\ldots, N \rbrace$ which corresponds to all the features in group $g$. A grouping of features may stem from any prior information about the data, for example which transcription factors tend to regulate the same genes or which genes belong to the same pathways, but also the grouping could be derived from the data itself, for example genes that show similar expression patterns and are grouped by clustering.
 
Here we assume the sparse-group setting, that is features are sparse on the between- and within-group level: in every group $g$ with corresponding features $\left(X_n\right)_{n \in \mathcal{G}^{-1}(g)}$, there are only a few variables with a pronounced influence on $y$ while the other variables from this group can be neglected (or there is no influence from group $g$ on $y$). This corresponds to the following regression scheme:
\begin{align*}
y_{m} = \sum_{g=1}^{G} \sum_{n\in \mathcal{G}^{-1}(g)} \beta_{n} \cdot x_{mn}+\varepsilon_{m}, \nlabel{eq:linsumgroup}
\end{align*}
where for every group $g$ some or none of the regression coefficients $\left(\beta_n\right)_{n \in \mathcal{G}^{-1}(g)}$ are different from zero. Of course, equation \eqref{eq:linsumgroup} is the same as equation \eqref{eq:linsum} but with a different ordering of the features, so it is up to the feature selection algorithm to make use of the grouping information.

We do not consider the intercept in the regression explicitly. There are two ways to deal with the intercept: either one centers the response $y$ around its mean ($y \leftarrow y - \bar{y}$) in the beginning, or one adds a column $X_0=(1, \ldots, 1)$ to the data matrix and a coefficient $\beta_0$ to the vector of coefficients ($X \leftarrow (X_0, X)$, $\beta \leftarrow (\beta_0, \beta)$), this way modeling the intercept as an additional feature (within its own group of size one).

 
In this work we extend the spike-and-slab approach to a sparse-group setting. The spike-and-slab is Bayesian variable selection method first introduced by \citet{mitchell_bayesian_1988} and subsequently \citet{geweke_variable_1994}, but in our work we follow the definition of \citet{george_approaches_1997} since it is more easily interpretable. A similar framework is formulated by \citet{kuo_variable_1998}.

\begin{figure}
\centering
\includegraphics[width=0.5\textwidth]{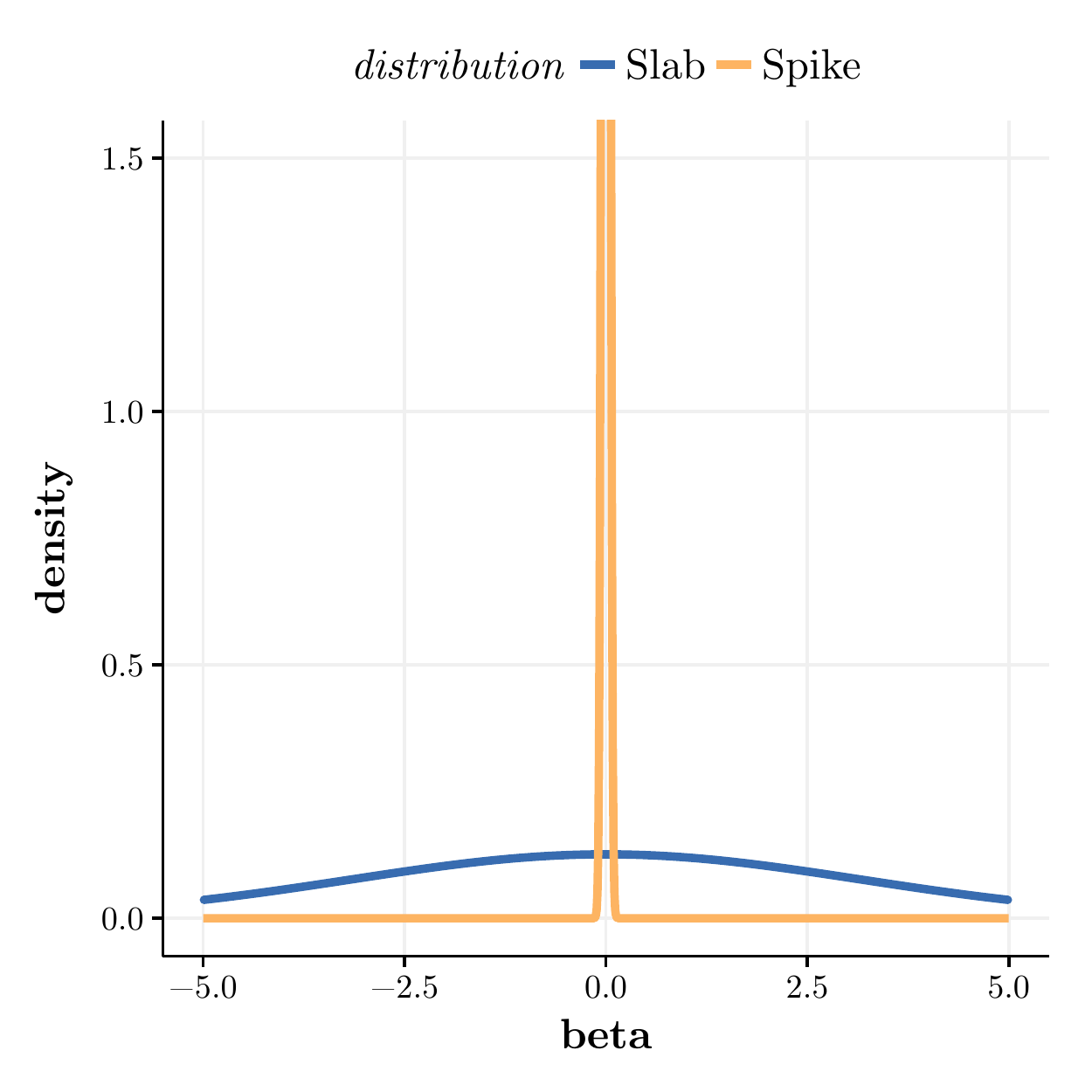}
\caption[Density function curves of spike and slab distributions.]{Density function curves of spike and slab distributions: both the spike and the slab are depicted as normal distributions $\mathcal{N}(\beta | 0, \sigma)$, with $\sigma_{\textnormal{slab}}=10$ and $\sigma_{\textnormal{spike}}=0.001$.}\label{fig_spikeslab}
\end{figure}

In general, the relationship between the observed $y$ and the predictors $X$ as well as $\beta$ and the error variance $\sigma_0$ can be described by the conditional probability
\begin{align*}
\mathbb{P}(y|\beta ,X) = \mathcal{N}(y|X\beta,\sigma_0^2 I).  \nlabel{eq:ynorm}
\end{align*}
Furthermore, if we assume that $\beta = (\beta_1, \ldots, \beta_N)$ is sparse and some $\beta_n$ are zero, we can introduce an auxiliary variable $Z = (Z_1, \ldots, Z_N) \in \lbrace 0, 1 \rbrace^{N}$ with $Z_n = 0$ indicating that $\beta_n = 0$ and from $Z_n = 1$ follows $\beta_n \neq 0$. We can give a prior distribution over $Z$:
\begin{align*}
\mathbb{P}(Z) = \prod_{n=1}^{N} \operatorname{Bern}(Z_{n} | p_{n}) = \prod_{n=1}^{N} p_n^{Z_n} \cdot (1-p_n)^{1-Z_n},  \nlabel{eq:zbern}
\end{align*}
where $p_n$ is the belief (probability) that the corresponding coefficient $\beta_n$ should be different from zero. This leads to a definition of the conditional probability $\mathbb{P}(\beta |Z)$ that is at the heart of the spike-and-slab approach:
\begin{align*}
\mathbb{P}(\beta |Z) = \prod_{n=1}^N \left( Z_{n} \cdot \mathcal{N}(\beta_{n}|0,\sigma_{\textnormal{slab}}^2) + (1-Z_{n}) \cdot \delta(\beta_{n}) \right).  \nlabel{eq:betamix}
\end{align*}
If $Z_n = 1$, the second summand disappears, and we draw $\beta_n$ from a normal distribution centered around zero but with a large variance parameter $\sigma_{\textnormal{slab}}$, thus $\beta_n \neq 0$. If $Z_n=0$, the first summand disappears, and we force $\beta_n$ to be equal to zero: $\delta(\cdot)$ is the distribution that puts probability mass $1$ to zero. Instead of $\delta(\cdot)$ one could also use a normal distribution $\mathcal{N}(\cdot|0,\sigma_{\textnormal{spike}}^2)$ with a small $\sigma_{\textnormal{spike}}$ \citep{george_approaches_1997}. For an illustration of the two distributions $\mathcal{N}(\cdot|0,\sigma_{\textnormal{slab}}^2)$ and $\delta(\cdot)$ see Figure \ref{fig_spikeslab}.

These representations (equations \ref{eq:ynorm}, \ref{eq:zbern} and \ref{eq:betamix}) allow for the following Bayesian framework to describe the linear regression scheme from equation \eqref{eq:linmatrix}:
\begin{align*}
\mathbb{P}(\beta,Z | y,X) = \frac{\mathbb{P}(y|\beta ,X)\cdot \mathbb{P}(\beta |Z) \cdot \mathbb{P}(Z)}{\mathbb{P}(y|X)}.  
\end{align*}
Given the data $(y, X)$, the task at hand is now to find the corresponding $\beta$ and $Z$ (and probabilities $p_n$, $n=1,\ldots,N$). One approach is to use Markov Chain Monte Carlo simulations respectively Gibbs sampling \citep[one example for this approach can be found in][]{george_approaches_1997}, but in this work we consider an alternative non-stochastic method called expectation propagation.

Note that we do not consider a prior on $\sigma_0$ in this work, which means this parameter needs to be specified beforehand.

We expand the Bayesian spike-and-slab framework to suit the regression scheme from equation \eqref{eq:linsumgroup}. To this end, we introduce another auxiliary variable $\Gamma = (\Gamma_{1},\ldots,\Gamma_{G}) \in \lbrace 0,1 \rbrace^G$ which handles the between-group sparsity while the $Z$ variable handles the within-group sparsity, given the value of $\Gamma$.

Given the data $(y,X)$ and a grouping $\mathcal{G}$ of the features, the following factorization of the true posterior distribution $\mathcal{P}(\beta, Z, \Gamma)=\mathbb{P}(\beta , Z , \Gamma| y,X)$ holds true (see also equation \ref{eq:Pfactored}):
\begin{align*}
\mathbb{P}(\beta , Z , \Gamma| y,X) & = \frac{\mathbb{P}(y|\beta ,X)\cdot \mathbb{P}(\beta |Z)\cdot \mathbb{P}(Z | \Gamma)\cdot \mathbb{P}(\Gamma)}{\mathbb{P}(y|X)} = \frac{1}{\mathbb{P}(y|X)} \prod_{i=1}^4 f_i(\beta,Z,\Gamma), \nlabel{eq:bayes} \\
\mathbb{P}(y|\beta ,X) &= f_1(\beta,Z,\Gamma) = \mathcal{N}(y|X\beta,\sigma_0^2 I), \nlabel{eq:normal} \\
\mathbb{P}(\beta |Z) &= f_2(\beta,Z, \Gamma) = \prod_{n=1}^{N} \left( Z_{n}\cdot \mathcal{N}(\beta_{n}|0,\sigma_{\textnormal{slab}}^2) + (1-Z_{n}) \cdot \delta(\beta_{n}) \right), \nlabel{eq:sas} \\
\mathbb{P}(Z | \Gamma) &= f_3(\beta,Z,\Gamma) = \prod_{n=1}^{N} \left( \Gamma_{\mathcal{G}(n)} \cdot \operatorname{Bern}(Z_n | p_n) + (1-\Gamma_{\mathcal{G}(n)}) \cdot \delta(Z_n) \right),\nlabel{eq:bern_within} \\
\mathbb{P}(\Gamma) &= f_4(\beta,Z,\Gamma) = \prod_{g=1}^{G} \operatorname{Bern}(\Gamma_{g} | \pi_{g}).\nlabel{eq:bern_between}
\end{align*}
The two-fold group sparsity assumption is realized by a refined spike-and-slab (equations \ref{eq:sas} and \ref{eq:bern_within}). If $\Gamma_{\mathcal{G}(n)} = 1$ for a certain $n$ \textbf{and} $Z_n = 1$, then the corresponding $\beta_{n}$ is different from zero, realized by $\beta_{n}$ drawn from the distribution $\mathcal{N}(0,\sigma_{\textnormal{slab}}^2)$ with a large $\sigma_{\textnormal{slab}}$, while if $\Gamma_{\mathcal{G}(n)} = 1$ but $Z_n=0$ \textbf{or} $\Gamma_{\mathcal{G}(n)}=0$ and subsequently $Z_n=0$ we have $\beta_{n} = 0$ with $\beta_{n}$ shrunken to zero by $\delta(\cdot)$. This way the variables $\Gamma_1, \ldots, \Gamma_G$ code for the between-group sparsity while for a fixed $g$ the variables $\left(Z_n\right)_{n\in\mathcal{G}^{-1}(g)}$ code for the within-group sparsity.

Given the data $(y, X)$ and the grouping of features $\mathcal{G}$, we want to find an estimate of the sparse coefficient vector $\beta$ (and probability vectors $p$ and $\pi$). To this end, we apply the algorithmic framework called expectation propagation.

\subsection{Expectation propagation algorithm}

Both the Bernoulli and the normal distribution are members of the class of exponential family distributions. This means that their respective probability densities can be written in the form
\begin{align*}
f(x) = h(x) \cdot g(\eta) \cdot \exp(\eta^T T(x)),
\end{align*}
where $\eta$ is called ``natural parameter'' and $T(x)$ is the ``sufficient statistic'' (see for example \citet[Chapter 10]{bishop_pattern_2006} for details on this and expectation propagation). More importantly, for all exponential family distributions the product (and to some extent the quotient) of two densities from an exponential family is again a density from an exponential family with updated parameters (up to a scaling factor depending on the parameters only). In our implementation we need these properties for the Bernoulli distribution and (multivariate) normal distribution.

The expectation propagation algorithm introduced by \citet{minka_expectation_2001} approximates a true (complicated) posterior distribution $\mathcal{P}$  with a (simpler) approximating distribution $\mathcal{Q}$ by iteratively minimizing the Kullback-Leibler divergence $\mathrm{KL}(\mathcal{P}||\mathcal{Q})$. As such, the expectation propagation algorithm is an example of deterministic approximate inference: the true posterior is too complex such that expectations are not analytically tractable since the resulting integrals do not have closed-form solutions and the dimensions of the parameter space prohibit numerical integration.

If $\mathcal{Q}$ is an exponential family distribution (and thus $\mathcal{Q}(x)=h(x) \cdot g(\eta) \cdot \exp(\eta^T T(x))$), the following holds true:
\begin{align*}
\mathbb{E}_{\mathcal{P}}\left[ T(x) \right] = \mathbb{E}_{\mathcal{Q}}\left[ T(x) \right],  \nlabel{eq:EPmatchgeneral}
\end{align*}
that is, to minimize the Kullback-Leibler divergence between $\mathcal{P}$ and $\mathcal{Q}$ we need to match the expectations under $\mathcal{P}$ and $\mathcal{Q}$ of the sufficient statistic $T$ of $\mathcal{Q}$.

The analytical matching under the complete distributions $\mathcal{P}$ and $\mathcal{Q}$ can be hard, but it gets easier if the distributions $\mathcal{P}$ and $\mathcal{Q}$ are factored, for example if $\mathcal{P}$ is a posterior distribution for some data $\mathcal{D}$:
\begin{align*}
\mathcal{P} &= \frac{1}{\mathbb{P}(\mathcal{D})} \prod_{i} f_i, \nlabel{eq:Pfactored}\\
\mathcal{Q} &= \frac{1}{\mathcal{Z}} \prod_i \tilde{f}_i,\nlabel{eq:Qfactored}
\end{align*}
where the factors $\tilde{f}_i$ belong to the exponential family of distributions, but they do not need to integrate to 1. That is why $\mathcal{Z}=\int\prod_i\tilde{f}_i$ is the normalization constant needed such that $\mathcal{Q}$ is a proper probability distribution which integrates to 1 (and $\mathcal{Z}$ approximates $\mathbb{P}(\mathcal{D})$).

Expectation propagation approximates $\mathrm{KL}(\mathcal{P}||\mathcal{Q})$ by first initializing the parameters of the $\tilde{f}_i$ and then cycling through the paired factors $f_i$ and $\tilde{f}_i$ one at a time and updating the corresponding parameters. Suppose we want to update factor $\tilde{f}_i$. First we remove $\tilde{f}_i$ from $\mathcal{Q}$, that is $\mathcal{Q}^{\backslash i} = \frac{\mathcal{Q}}{\tilde{f}_i}$, where $\mathcal{Q}^{\backslash i}$ is a probability distribution up to a normalization constant. Now we want to update $\tilde{f}_i \rightarrow \tilde{f}^{\textnormal{new}}_i$ such that $\tilde{f}^{\textnormal{new}}_i \cdot \mathcal{Q}^{\backslash i}$ is close to $f_i \cdot \mathcal{Q}^{\backslash i}$, or, with $\mathcal{Q}^{\textnormal{new}}$ the normalized version of $\tilde{f}^{\textnormal{new}}_i \cdot \mathcal{Q}^{\backslash i}$, we minimize $\mathrm{KL}(\frac{1}{\mathcal{Z}_i}f_i \cdot \mathcal{Q}^{\backslash i}||\mathcal{Q}^{\textnormal{new}})$ for some (unimportant) normalization constant $\mathcal{Z}_i$. Equation \eqref{eq:EPmatchgeneral} tells us we need to match the sufficient statistics of these two distributions, where calculus and arithmetic yield the updated parameters of $\mathcal{Q}^{\textnormal{new}}$. After we have found $\mathcal{Q}^{\textnormal{new}}$ we get the parameters of our updated $\tilde{f}^{\textnormal{new}}_i$ from $\tilde{f}^{\textnormal{new}}_i = \mathcal{Z}_i \frac{\mathcal{Q}^{\textnormal{new}}}{\mathcal{Q}^{\backslash i}}$, where the value of $\mathcal{Z}_i$ is not needed to update the parameters.


In our implementation of the expectation propagation algorithm, the procedure is based on the following approximation between the factorized true posterior distribution $\mathcal{P}$ and the likewise factorized approximate posterior distribution $\mathcal{Q}$ (see equations \ref{eq:Pfactored} and \ref{eq:Qfactored}):
\begin{align*}
\mathcal{P}(\beta,Z,\Gamma) &\approx \mathcal{Q}(\beta,Z,\Gamma), \\
\textnormal{where } \mathcal{P}(\beta,Z,\Gamma) &=  \frac{1}{\mathbb{P}(y|X)} \prod_{i=1}^4 f_i(\beta,Z,\Gamma) \textnormal{ (see Eq. \eqref{eq:bayes}),} \nlabel{eq:myp} \\
\textnormal{and } \mathcal{Q}(\beta,Z,\Gamma) &= \frac{1}{\mathcal{Z}} \cdot \prod_{i=1}^4 \tilde{f}_i(\beta,Z,\Gamma), \nlabel{eq:myq}\\
\textnormal{with } \tilde{f}_1(\beta,Z,\Gamma) &= \tilde{s}_1 \cdot \mathcal{N}(\beta|\tilde{m}_1,\tilde{V}_1), \nlabel{eq:f1}\\
\tilde{f}_2(\beta,Z,\Gamma) &= \tilde{s}_2 \cdot \prod_{n=1}^{N} \mathcal{N}(\beta_n|\tilde{m}_{2,n},\tilde{V}_{2,n}) \cdot \operatorname{Bern}(Z_n | \tilde{p}_{2,n}),\nlabel{eq:f2}\\
\tilde{f}_3(\beta,Z,\Gamma) &= \tilde{s}_3 \cdot \prod_{n=1}^N \operatorname{Bern}(Z_n | \tilde{p}_{3,n}) \cdot \operatorname{Bern}(\Gamma_{g(n)} | \tilde{\pi}_{3,n}), \nlabel{eq:f3}\\
\tilde{f}_4(\beta,Z,\Gamma) &= \tilde{s}_4 \cdot \prod_{g=1}^{G} \operatorname{Bern}(\Gamma_{g} | \tilde{\pi}_{4,g}), \nlabel{eq:f4}\\
\textnormal{and thus } \mathcal{Q}(\beta,Z,\Gamma) &= \mathcal{N}(\beta|\tilde{m},\tilde{V}) \cdot \prod_{g=1}^G \operatorname{Bern}(\Gamma_g | \tilde{\pi}_{g}) \cdot \prod_{n=1}^{N} \operatorname{Bern}(Z_{n} | \tilde{p}_{n}) \nlabel{eq:qdistr}.
\end{align*}
The parameters $\tilde{s}_1$ to $\tilde{s}_4$ ensure that $\tilde{f}_i\mathcal{Q}^{\backslash i}$ and $f_i\mathcal{Q}^{\backslash i}$ integrate up to the same value.
The particular form of $\mathcal{Q}$ in equation \eqref{eq:qdistr} as a product of the terms in equations \eqref{eq:f1} to \eqref{eq:f4} (divided by the normalization factor $\mathcal{Z}$) follows from the above mentioned properties of exponential family distributions.

The expectation propagation algorithm starts with an initial guess for the parameters $\tilde{m}$, $\tilde{V}$, $\tilde{p}$ and $\tilde{\pi}$ of $\mathcal{Q}$ and the parameters of the functions $\tilde{f}_i$ and iteratively matches the expectations under $\mathcal{Q}$ and $\mathcal{P}$ of the sufficient statistics of $\mathcal{Q}$ until convergence in the parameters is reached:
\begin{align*}
\mathbb{E}_{\mathcal{Q}} \left[ (\beta, \beta\beta^T, Z, \Gamma)^T \right] \stackrel{!}{=} \mathbb{E}_{\mathcal{P}} \left[ (\beta, \beta\beta^T, Z, \Gamma)^T \right]. \nlabel{eq:EPmatchDogss}
\end{align*}

The sufficient statistic of a normal distribution $\mathcal{N}(x|\mu,\sigma^2)$ is $(x, x^2)$, while the sufficient statistic of a ($N$-dimensional) multivariate normal distribution $\mathcal{N}(x|\mu,\Sigma)$ is $(x, xx^T)$ with $x\in\mathbb{R}^N$ and $xx^T\in\mathbb{R}^{N\times N}$. The sufficient statistic of a Bernoulli distribution $\operatorname{Bern}(x | p)$ is just $x$.

Since $\mathcal{P}$ and $\mathcal{Q}$ are factored (equations \ref{eq:myp} and \ref{eq:myq}), we update the parameters in expectation propagation by taking turns updating $\tilde{f}_1$, $\tilde{f}_2$, $\tilde{f}_3$ and $\tilde{f}_4$. The update operations are derived in the next section.

\subsection{Expectation propagation update operations}\label{sec:epupdates}

We need to initialize all parameters of the respective normal and Bernoulli distributions before the start of the expectation propagation algorithm. It is an advantage of a Bayesian framework that one can include prior information into these initial parameters and give the algorithm a ``head start''. This is especially useful for the prior probabilities for feature inclusion $p_0$ and\slash or group inclusion $\pi_0$. If initial probabilities are not provided, we set the default parameters $p_{0,n}=\pi_{0,g}=\frac{1}{2}$, $n=1,\ldots,N$, $g=1,\ldots,G$. Also, the parameter $\sigma_0$ for the random noise needs to be chosen. \citet{george_approaches_1997} include the estimation of $\sigma_0$ by adding another prior for this parameter that then needs to be updated along with the other probabilities (they use an inverse-gamma prior distribution). In this work, to reduce the mathematical and numerical overhead, we initialize this parameter in the beginning and keep it untouched during the algorithm. Our simulations do not indicate that using an inverse-gamma prior distribution is advantageous (see Section \ref{sec:noise}). Finally we need to initialize the parameter $\sigma_{\textnormal{slab}}$, we discuss and analyze this parameter in Section \ref{sec:slabparameter}. 

An important aspect is numerical stability. Since we multiply many (small) probabilities, we do not use the probability parameters of the Bernoulli distributions directly, but instead their $\operatorname{logit}$-transformed versions: $r = \operatorname{logit}(p)=\log\frac{p}{1-p}$ and $\varrho = \operatorname{logit}(\pi)$, as was done by \citet{hernandez-lobato_balancing_2010}. Further numerical issues are discussed at the end of Section \ref{sec:iterative}.

In the expectation propagation framework, more precisely the factorizations in equations \eqref{eq:Pfactored} and \eqref{eq:Qfactored}, we do not consider the factored distributions $\tilde{f}_i$, $i=1,\ldots, 4$ directly, but rather we use the fully factorized \citep[Section 10.7.2]{bishop_pattern_2006} feature specific distributions $\tilde{f}_{i,n}$, $i=1,\ldots, 4$, $n=1,\ldots,N$. In the case of $\tilde{f}_{2,n}$ this translates to $\tilde{V}_2$ being a diagonal matrix (instead of a dense matrix) and the feature-wise variances are updated independently of each other. We capture the dependent effects with the initialization of the $\tilde{V}_1$ matrix and those effects are translated into the $\tilde{V}$ parameter matrix in each iteration of the algorithm. Thus we do not match the expectations under $\mathcal{Q}$ and $\mathcal{P}$ of the original sufficient statistic $(\beta, \beta\beta^T,Z,\Gamma)$, but rather element-wise $(\beta_n, \beta_n^2, Z_n, \Gamma_n)$ separately.

\subsubsection{Initialization}

The values of the $\tilde{f}_1$ parameters are easy to obtain and do not need to be re-estimated by the algorithm since $\tilde{f}_1$ and $f_1$ have the same form (a multivariate normal distribution), thus we derive the exact approximations for $\tilde{m}_1$ and $\tilde{V}_1$ directly from the ordinary least squares estimate $\beta \approx \hat{\beta} = (X^TX)^{-1}X^Ty$. It is more convenient to save $\tilde{V}_1^{-1}$ and $\tilde{V}_1^{-1}\tilde{m}_1$ for the remaining operations of the algorithm. Also, $\tilde{V}_1^{-1}$ might not be of full rank and thus $\tilde{V}_1$ might not exist and in turn $\tilde{m}_1$ would not be unique. Thus, the initial and final estimates for the parameters of $\tilde{f}_1$ are given by:
\begin{align*}
\tilde{V}_1^{-1} &= \frac{1}{\sigma_0^2}X^TX, \\
\tilde{V}_1^{-1}\tilde{m}_1 &= \frac{1}{\sigma_0^2}X^Ty.
\end{align*}
The initial guess for the variance $\tilde{V}_{2,n}$ in $\tilde{f}_{2,n}$ is the prior probability of choosing feature $n$ multiplied by the variance of the slab. The initial estimate for $\tilde{m}_{2,n}$ is the non-informative $\vec{0}$ and the initial values for $\tilde{p}_{2,n}$ as well as $\tilde{p}_{3,n}$ and $\tilde{\pi}_{3,n}$ will default to $p_0$ and $\pi_0$ (more precisely, their $\operatorname{logit}$ versions):
\begin{align*}
\tilde{V}_2 &= \sigma_{\textnormal{slab}}^2\cdot p_0,\\
\tilde{V}_2^{-1}\tilde{m}_2 &= \vec{0}, \\
\tilde{r}_2 &= \tilde{r}_3 = r_0, \\
\tilde{\varrho}_3 &= \varrho_0.
\end{align*}
For the parameters of $\tilde{f}_4$ we have the same situation like $\tilde{f}_1$: both $\tilde{f}_4$ and $f_4$ have the same form (a product of $G$ Bernoulli distributions) and thus we set $\tilde{\pi}_{4}$ equal to $\pi_0$ (respectively the $\operatorname{logit}$ counterparts) and do not touch these parameters for the rest of the algorithm:
\begin{align*}
\tilde{\varrho}_4 = \varrho_0.
\end{align*}
With all these parameters of the factor distributions initialized we can finally initialize the parameters of the product distribution $\mathcal{Q}$, too. This is done by using the properties of the product of normal and Bernoulli distributions, see Appendix B. Thus the initialization for $\mathcal{Q}$ is given by:
\begin{align*}
\tilde{V} &= \left( \tilde{V}_1^{-1} + \tilde{V}_2^{-1} \right)^{-1}, \\
\tilde{m} &= \tilde{V} \left( \tilde{V}_1^{-1}\tilde{m}_1 + \tilde{V}_2^{-1}\tilde{m}_2 \right), \\
\tilde{r} &= r_0, \\
\tilde{\varrho} &= \varrho_0.
\end{align*}

\subsubsection{Iterative updates}\label{sec:iterative}

After initializing all the parameters we update iteratively the parameters of the $\tilde{f}_i$ distributions. We give the explicit iterative updates in the following, the derivations of these updates are found in Appendix A. 


The expectation propagation algorithm matches expectations under $\mathcal{P}$ and $\mathcal{Q}$ of the sufficient statistics of $\mathcal{Q}$ (equation \ref{eq:EPmatchDogss}) by iteratively minimizing the Kullback-Leibler divergence $\operatorname{KL}(\mathcal{P}||\mathcal{Q})$. Since $\mathcal{P}$ and $\mathcal{Q}$ are factored (equations \ref{eq:myp} and \ref{eq:myq}), this is done by iterating through the factors $\tilde{f}_i$, $i=1, \ldots, 4$ and updating their respective parameters and $\mathcal{Q}$ in turns. In fact, only $i=2,3$ are considered, since $f_1$ and $\tilde{f}_1$ respectively $f_4$ and $\tilde{f}_4$ have the same form by choice. As such, the values for $\tilde{V}_1$ and $\tilde{m}_1$ (respectively $\tilde{V}_1^{-1}$ and $\tilde{V}_1^{-1}\tilde{m}_1$) as well as $\tilde{\varrho}_4$ do not need to be updated,  and thus their respective initial and final values are given by:
\begin{align*}
\tilde{V}_1^{-1} &= \frac{1}{\sigma_0^2}X^TX, \\
\tilde{V}_1^{-1}\tilde{m}_1 &= \frac{1}{\sigma_0^2}X^Ty,\\
\tilde{\varrho}_4 &= \varrho_0.
\end{align*}
This leaves us with the updates for $\tilde{f}_2$ and $\tilde{f}_3$. First, we cycle through the $\tilde{f}_{2,n}$ and find the parameters of the updated $\tilde{f}_{2,n}^{\textnormal{new}}$ via finding $\mathcal{Q}^{\backslash 2,n}$, then updating $\mathcal{Q}$ with the rules for the product of exponential family distributions. Second, the same is repeated for $\tilde{f}_{3}$ by cycling through the factors $\tilde{f}_{3,n}$ and finding the parameters of the updated $\tilde{f}_{3,n}^{\textnormal{new}}$ via finding $\mathcal{Q}^{\backslash 3,n}$ and afterwards updating $\mathcal{Q}$ like before.

Updates for $\tilde{f}_{2,n}$: The parameters $\tilde{V}_{n}^{\backslash 2,n}$, $\tilde{m}_{n}^{\backslash 2,n}$ and $\tilde{r}_n^{\backslash 2,n}$ of $\mathcal{Q}^{\backslash 2,n} \propto \mathcal{Q} \slash \tilde{f}_{2,n}$ are derived by using the rules for quotients of Bernoulli or normal distributions (Appendix B):
\begin{align*}
\tilde{V}_{n}^{\backslash 2,n} &= \left( \tilde{V}_{nn}^{-1} - \tilde{V}_{2,n}^{-1} \right)^{-1}, \\
\tilde{m}_{n}^{\backslash 2,n} &= \tilde{V}_{n}^{\backslash 2,n} \cdot \left( \tilde{V}_{nn}^{-1} \cdot \tilde{m}_n - \tilde{V}_{2,n}^{-1} \cdot \tilde{m}_{2,n} \right), \\
\tilde{r}_n^{\backslash 2,n} &= \tilde{r}_n - \tilde{r}_{2,n}.
\end{align*}
We find the updated $\tilde{f}_{2,n}^{\textnormal{new}}$ by minimizing the Kullback-Leibler divergence between $f_{2,n}\cdot \mathcal{Q}^{\backslash 2,n}$ and $\tilde{f}_{2,n}^{\textnormal{new}}\cdot \mathcal{Q}^{\backslash 2,n}$. The updated parameters of $\tilde{f}_{2,n}^{\textnormal{new}}$ are given by (derivation in the Appendix A):
\begin{align*}
\tilde{r}_{2,n}^{\textnormal{new}} &= \frac{1}{2} \cdot \left( \log \left( \frac{\tilde{V}_{n}^{\backslash 2,n}}{\tilde{V}_{n}^{\backslash 2,n}+\sigma^2_{\textnormal{slab}}} \right) + (\tilde{m}_{n}^{\backslash 2,n})^2 \cdot \left( 1\slash \tilde{V}_{n}^{\backslash 2,n} - 1\slash(\tilde{V}_{n}^{\backslash 2,n}+\sigma^2_{\textnormal{slab}}) \right) \right),\\
\tilde{V}_{2,n}^{\textnormal{new}} &= \frac{1}{a_n^2-b_n} - \tilde{V}_{n}^{\backslash 2,n}, \\
\tilde{m}_{2,n}^{\textnormal{new}} &= \tilde{m}_{n}^{\backslash 2,n} - a_n\cdot (\tilde{V}_{2,n}^{\textnormal{new}} + \tilde{V}_{n}^{\backslash 2,n}),\\
\textnormal{with } a_n &= p^{\textnormal{aux}}_{n} \cdot \frac{\tilde{m}_{n}^{\backslash 2,n}}{\tilde{V}_{n}^{\backslash 2,n}+\sigma^2_{\textnormal{slab}}} + (1-p^{\textnormal{aux}}_{n}) \cdot \frac{\tilde{m}_{n}^{\backslash 2,n}}{\tilde{V}_{n}^{\backslash 2,n}}, \\
b_n &= p^{\textnormal{aux}}_{n} \cdot \frac{(\tilde{m}_{n}^{\backslash 2,n})^2-\tilde{V}_{n}^{\backslash 2,n}-\sigma^2_{\textnormal{slab}}}{(\tilde{V}_{n}^{\backslash 2,n}+\sigma^2_{\textnormal{slab}})^2} + (1-p^{\textnormal{aux}}_{n}) \cdot \frac{(\tilde{m}_{n}^{\backslash 2,n})^2-\tilde{V}_{n}^{\backslash 2,n}}{(\tilde{V}_{n}^{\backslash 2,n})^2}\\
\textnormal{and } p^{\textnormal{aux}}_{n} &= \operatorname{sigmoid}(\tilde{r}_{2,n}^{\textnormal{new}}+\tilde{r}_n^{\backslash 2,n}),
\end{align*}
where the sigmoid function is the inverse function of the logit function: $\operatorname{sigmoid}(r)=1\slash(1+\exp(-r))=\exp(r)\slash(1+\exp(r))$.

The updates for $\mathcal{Q}$ after updating $\tilde{f}_{2,n}^{\textnormal{new}}$ are derived from the rules for the product of Bernoulli and normal distributions (Appendix B):
\begin{align*}
\tilde{V} &= \left( \tilde{V}_1^{-1} + \left(\tilde{V}_2^{\textnormal{new}}\right)^{-1} \right)^{-1}, \nlabel{eq:bottleneck} \\
\tilde{m} &= \tilde{V} \left( \tilde{V}_1^{-1}\tilde{m}_1 + \left(\tilde{V}_2^{\textnormal{new}}\right)^{-1} \tilde{m}_{2}^{\textnormal{new}} \right), \\
\tilde{r}_n &= \tilde{r}_{2,n}^{\textnormal{new}} + \tilde{r}_{3,n},\\
\tilde{\varrho} &\textnormal{ does not change.}
\end{align*}
Updates for $\tilde{f}_{3,n}$: The parameters $\tilde{r}_n^{\backslash 3,n}$ and $\tilde{\varrho}_n^{\backslash 3,n}$ of $\mathcal{Q}^{\backslash 3,n} \propto \mathcal{Q} \slash \tilde{f}_{3,n}$ are derived by using the rules for quotients of Bernoulli distributions:
\begin{align*}
\tilde{\varrho}_{n}^{\backslash 3,n} &= \tilde{\varrho}_{\mathcal{G}(n)} - \tilde{\varrho}_{3,n},\\
\tilde{r}_{n}^{\backslash 3,n} &= \tilde{r}_n - \tilde{r}_{3,n}.
\end{align*}
We find the updated $\tilde{f}_{3,n}^{\textnormal{new}}$ by minimizing the Kullback-Leibler divergence between $f_{3,n}\cdot \mathcal{Q}^{\backslash 3,n}$ and $\tilde{f}_{3,n}^{\textnormal{new}}\cdot \mathcal{Q}^{\backslash 3,n}$. The final analytical parameter updates of $\tilde{\varrho}_{3,n}^{\textnormal{new}}$ and $\tilde{r}_{3,n}^{\textnormal{new}}$ are given by: 
\begin{align*}
\tilde{\varrho}_{3,n}^{\textnormal{new}} &= -\log(1-\tilde{p}_{n}^{\backslash 3,n}) + \log(\tilde{p}_{n}^{\backslash 3,n}\cdot p_{0,n} + (1-\tilde{p}_{n}^{\backslash 3,n}) \cdot (1-p_{0,n}))\\
&= \log(1+p_{0,n}\cdot(\exp(\tilde{r}_{n}^{\backslash 3,n})-1)), \\
\tilde{r}_{3,n}^{\textnormal{new}} &= \operatorname{logit}(\tilde{\pi}_{n}^{\backslash 3,n} \cdot p_{0,n}) \\
&= \log p_{0,n} - \log(1-p_{0,n}+\exp(-\tilde{\varrho}_{n}^{\backslash 3,n})).
\end{align*}
The updates for $\mathcal{Q}$ after updating $\tilde{f}_{3,n}$ are derived from the rules for the product of Bernoulli distributions (Appendix B):
\begin{align*}
\tilde{V} &\textnormal{ does not change,} \\
\tilde{m} &\textnormal{ does not change,} \\
\tilde{\varrho}_{\mathcal{G}(n)} &= \tilde{\varrho}_{4,\mathcal{G}(n)} + \sum_{l:\mathcal{G}(l)=\mathcal{G}(n)} \tilde{\varrho}_{3,l}^{\textnormal{new}}, \\
\tilde{r}_n &= \tilde{r}_{2,n} + \tilde{r}_{3,n}^{\textnormal{new}}.
\end{align*}

If $p_{0,n}=0.5$, the update operations for $\tilde{\varrho}_{3,n}^{\textnormal{new}}$ and $\tilde{r}_{3,n}^{\textnormal{new}}$ are given by:
\begin{align*}
\tilde{\varrho}_{3,n}^{\textnormal{new}} &= \log(0.5) + \log(1+\exp(\tilde{r}_{n}^{\backslash 3,n})), \\
\tilde{r}_{3,n}^{\textnormal{new}} &= - \log(1+2\exp(-\tilde{\varrho}_{n}^{\backslash 3,n})).
\end{align*}

Thus, the (reasonable) choice of $p_{0,n}=0.5$ is numerically advantageous, too, since it allows for the efficient use of \texttt{log1p} and \texttt{logsumexp} functions in the implementation of the update operations.

Note that the calculation of $\tilde{V}$ in equation \eqref{eq:bottleneck} is the bottleneck of the algorithm's complexity. In the case of $N>M$ (more features than observations) there is a more efficient way to invert the matrix from equation \eqref{eq:bottleneck}, that is the Woodbury formula \citep{hager_updating_1989}:
\begin{align*}
\tilde{V} &= \left( \tilde{V}_1^{-1} + \tilde{V}_2^{-1} \right)^{-1} \\
&= \tilde{V}_2 - \tilde{V}_2 X^T \left( \sigma^2_0 I + X \tilde{V}_2 X^T \right)^{-1} X \tilde{V}_2.
\end{align*}
Instead of inverting an ($N\times N$)-matrix, because of the choice of $\tilde{V}_1^{-1}=\frac{1}{\sigma_0^2}X^TX$ the Woodbury matrix identity allows us to invert an ($M\times M$)-matrix instead.

Rarely the variance $\tilde{V}_{n}^{\backslash 2,n}$ might be negative before the update operation, in this case we do not perform an update operation on $\tilde{f}_{2,n}$. In addition, the variance $\tilde{V}_{2,n}^{\textnormal{new}}$ might be negative after the corresponding update operation, which is a well known problem \citep{minka_family_2001}. This arises as a compensation for errors in the first factor $\tilde{f}_{1}$, but hampers the ability of the expectation propagation algorithm to converge \citep{seeger_bayesian_2008}. To improve convergence, we follow the observations of \citep{hernandez-lobato_balancing_2010} and apply the constraint of $\tilde{V}_{2,n}^{\textnormal{new}}>0$, such that we replace its value by a large constant (100) whenever it turns out to be negative. 

Minimizing the Kullback-Leibler divergence is an optimization problem with a single global optimum which can be found by matching the sufficient statistics as described above. The expectation propagation algorithm is not guaranteed to converge to this global solution, but often converges to a fixed point \citep{minka_expectation_2001}. Furthermore, \citet{minka_expectation-propagation_2002} introduce damping of the updated factors to secure convergence of expectation propagation:
\begin{align*}
\tilde{f}_{i,n} = \left( \tilde{f}_{i,n}^{\textnormal{new}} \right)^{\alpha} \cdot \left( \tilde{f}_{i,n}^{\textnormal{old}} \right)^{1-\alpha}, \textnormal{ where } \alpha \in [0,1].
\end{align*}
By setting $\alpha = 0.9$ and decaying it by $1\%$ in every step of the algorithm we follow the advice of \citet{hernandez-lobato_generalized_2013}. The updated parameters $(\tilde{V}_{2,n}^{\textnormal{new}})^{-1}$, $(\tilde{V}_{2,n}^{\textnormal{new}})^{-1}\tilde{m}_{2,n}^{\textnormal{new}}$, $\tilde{r}_{2,n}^{\textnormal{new}}$, $\tilde{r}_{3,n}^{\textnormal{new}}$ and $\tilde{\varrho}_{3,n}^{\textnormal{new}}$ before updating the parameters of $\mathcal{Q}$ are then given by:
\begin{align*}
(\tilde{V}_{2,n}^{\textnormal{new}})^{-1} &\leftarrow \alpha \cdot (\tilde{V}_{2,n}^{\textnormal{new}})^{-1} + (1-\alpha) \cdot (\tilde{V}_{2,n}^{\textnormal{old}})^{-1},\\
(\tilde{V}_{2,n}^{\textnormal{new}})^{-1}\tilde{m}_{2,n}^{\textnormal{new}} &\leftarrow \alpha \cdot (\tilde{V}_{2,n}^{\textnormal{new}})^{-1}\tilde{m}_{2,n}^{\textnormal{new}} + (1-\alpha)\cdot (\tilde{V}_{2,n}^{\textnormal{old}})^{-1}\tilde{m}_{2,n}^{\textnormal{old}},\\
\tilde{r}_{2,n}^{\textnormal{new}} &\leftarrow \alpha\cdot \tilde{r}_{2,n}^{\textnormal{new}} + (1-\alpha)\cdot \tilde{r}_{2,n}^{\textnormal{old}},\\
\tilde{r}_{3,n}^{\textnormal{new}} &\leftarrow \alpha\cdot \tilde{r}_{3,n}^{\textnormal{new}} + (1-\alpha)\cdot \tilde{r}_{3,n}^{\textnormal{old}},\\ \tilde{\varrho}_{3,n}^{\textnormal{new}} &\leftarrow \alpha\cdot \tilde{\varrho}_{3,n}^{\textnormal{new}} + (1-\alpha)\cdot \tilde{\varrho}_{3,n}^{\textnormal{old}}.
\end{align*}

Our implementation of the method described above is available as a package \linebreak\mbox{(\url{https://github.com/edgarst/dogss})} within the statistical programming language and environment R \citep{r_core_team_r:_2017}. 

\section{Signal Recovery}

Here we will test the new method on simulated data $y=X\beta+\varepsilon$ with known $\beta$ and noise $\varepsilon$, where we try to recover $\beta$ from the observations $y$ and $X$ to make predictions on held-out data and measure computing time.

We compare six different methods:
\begin{enumerate}[(i)]
\item \texttt{dogss}: Our implementation of our new method, the sparse-group Bayesian feature selection with expectation propagation. \texttt{dogss} is an abbreviation for \textbf{do}uble \textbf{g}roup-sparse \textbf{s}pike-and-\textbf{s}lab.
\item \texttt{ssep}: Our implementation (as a special case of \texttt{dogss} without grouping of features) of the standard \textbf{s}pike-and-\textbf{s}lab with \textbf{e}xpectation \textbf{p}ropagation.
\item \texttt{sgl}: The sparse-group lasso from \citet{simon_sparse-group_2013}, implemented within their R package \texttt{SGL}.
\item \texttt{gglasso}: The group lasso from \citet{yuan_model_2006}, implemented by \citet{yang_fast_2015} within the R package \texttt{gglasso}.
\item \texttt{lasso}: The standard lasso from \citet{tibshirani_regression_1996}, implemented by \citet{friedman_regularization_2010} within the R package \texttt{glmnet}.
\item \texttt{bsgsss}: An implementation of a different sparse-group Bayesian feature selection with Gibbs sampling from \citet{xu_bayesian_2015}, implemented by \citet{liquet_bayesian_2017} within the R package \texttt{MBSGS}.
\end{enumerate}
All simulations, implementations, top-level method calls and calculations for evaluation of the results were done in R, a statistical programming language and environment \citep{r_core_team_r:_2017}.


If a ranking of retrieved coefficients of the features is provided (which is the case for all methods considered), one can measure the performance of a method in respect to correctly identified non-zero coefficients (without considering the actual value of the coefficients) along the ranking (parameterized for example by some value $\lambda$). This is done by evaluating the receiver operating characteristics (ROC) and precision-recall (PR) curves. The ROC curve plots the false-positive rate ($\operatorname{FPR}$ or 1-specificity) against the true-positive rate ($\operatorname{TPR}$, sensitivity or recall), while the PR curve plots the $\operatorname{TPR}$ against the precision ($\operatorname{Prec}$), where
\begin{align*}
\operatorname{TPR}(\lambda) = \frac{\operatorname{TP}(\lambda)}{k},\ \operatorname{FPR}(\lambda) = \frac{\operatorname{FP}(\lambda)}{N^\ast-k},\ \operatorname{Prec}(\lambda) = \frac{\operatorname{TP}(\lambda)}{\operatorname{TP}(\lambda)+\operatorname{FP}(\lambda)},
\end{align*}
and $N^\ast$ is here the number of all features for the problem of signal recovery ($N^\ast=N$) or the number of all possible edges $N^\ast=(P^2-P)\slash2$ in the undirected network graph of $P$ nodes in Section 4. To do this analysis, a gold standard of correct labels (existent/absent edges) is needed which is compared to the algorithm's retrieved edges. 


We can assess the overall-performance of a method by calculating the area under the (ROC or PR) curve, or $\operatorname{AUROC}$ respectively $\operatorname{AUPR}$. An estimator that chooses positive and negative labels randomly has $\operatorname{AUROC}=0.5$ and $\operatorname{AUPR}=k\slash N^\ast$.

The ranking of the recovered features for the Bayesian approaches (\texttt{dogss}, \texttt{ssep} and \texttt{bsgsss}) is done via the probabilities associated with every feature. The ranking of features for the lasso methods (\texttt{lasso}, \texttt{gglasso} and $\texttt{sgl}$) is along the penalty parameter $\lambda$. 



We also compare running times of the algorithms for the problem of signal recovery, and to this end we set the machine precision tolerance of all methods to the same value ($10^{-5}=0.00001$) and allowed a maximum of 1000 iterations each. On top of this we provided the same $\lambda$ sequence (of 100 $\lambda$ values) to all lasso methods, which is the one calculated by the implementation within the \texttt{glmnet} package of the standard lasso.

As a complement to the ROC/PR curve analysis, we evaluate the prediction errors of the different methods. For every scenario, we generate $100$ additional observations as a test set. We calculate relative prediction errors on the test set using the retrieved parameters from the training data. In signal recovery, this corresponds to the following relative residual sum of squared errors $E$ for test data $y_{\textnormal{test}}=(y_m)_{m=1}^{100}$ and $X_{\textnormal{test}}=(x_{mn})$, $m=1,\ldots,100$, $n=1,\ldots,N$:
\begin{align*}
E=\frac{\sum_{m=1}^{100} (y_m - \sum_{n=1}^N\hat{\beta}_n x_{mn})^2}{\sum_{m=1}^{100} y_m^2}.
\end{align*}

For signal recovery, we generate $M$ random observations of $N$ features, each drawn from a normal distribution with mean 0 and variance 1, this gives the data matrix $X$. For now, the features are drawn independently of each other. The features are divided into $G$ groups by sampling independently for every feature a group index from $\lbrace 1, \ldots, G \rbrace$, thus we have different numbers of features $(N_1, \ldots, N_G)$ in every group. The resulting $(M\times N)$-matrix $X$ of observations is multiplied with a $k$-sparse coefficient vector $\beta$ of length $N$ ($k$ out of the $N$ coefficients are different from zero, drawn independently from a uniform distribution on $[-5,5]$). The $k$ non-zero coefficients are only chosen within 3 random groups, thus we have sparsity on the group level with 3 non-zero groups and $G-3$ groups with all zero coefficients. Finally we add some noise $\varepsilon \sim \mathcal{N}(0, \sigma_0^2)$ and obtain the response vector $y=X\beta + \varepsilon$.

\begin{figure}
\centering
\includegraphics[width=\textwidth]{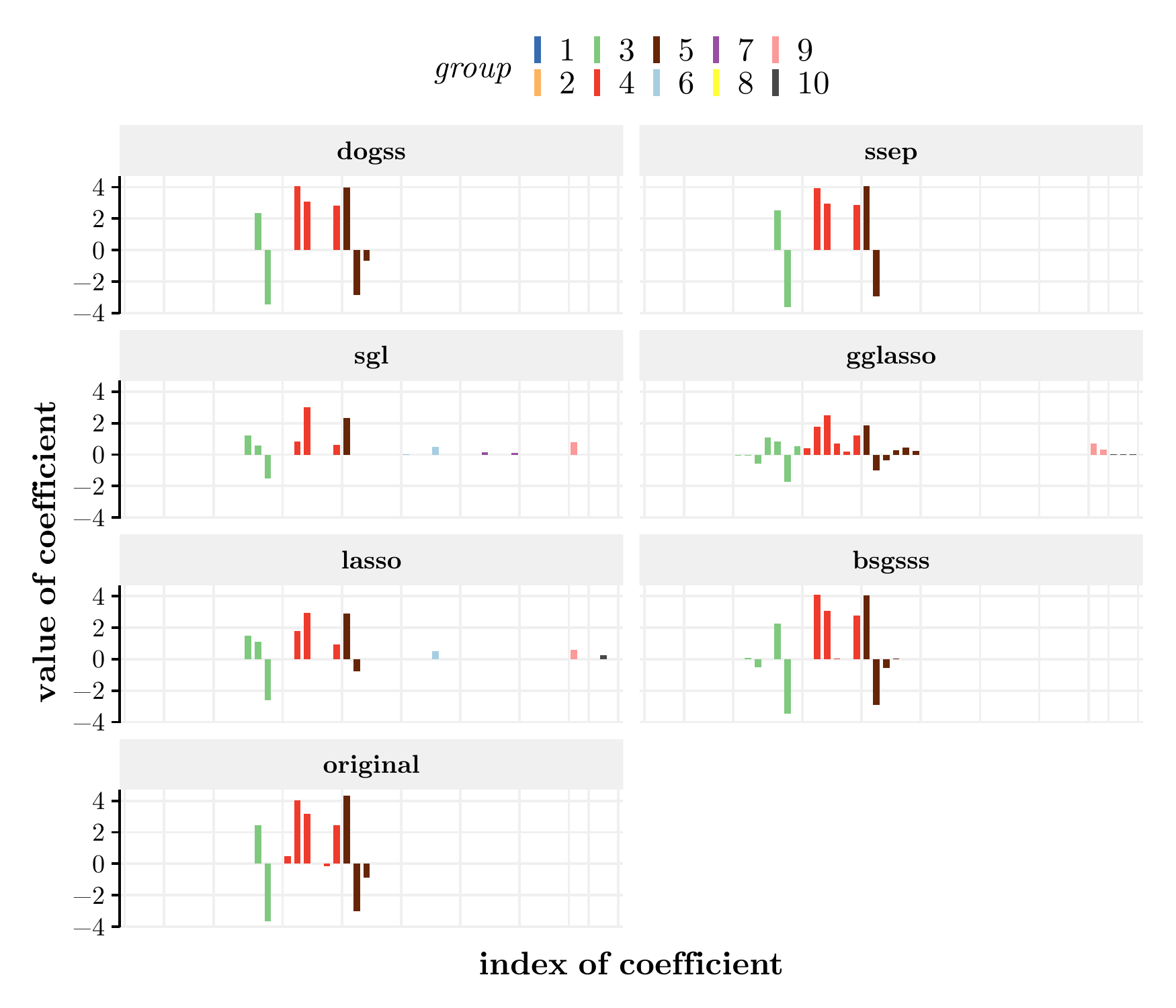}
\caption[Needle plot showing retrieved regression coefficients (signal recovery).]{Needle plot for signal recovery: results from one simulation with parameters $(M,N,G,k,\sigma_0)=(30,50,10,10,1)$. Features are aligned by indices along the x-axis, the height of bars corresponds to the value of the coefficient. The ``original'' box shows the true signal, while the other six boxes show the retrieved coefficients from six different methods.}\label{fig_signalrecovery}
\end{figure}

We feed the matrix $X$, the response $y$ and the group indices into the different algorithms and compare the performance on the resulting estimates $\hat{\beta}$ of the vector of coefficients. See an example of the reconstructed coefficient vectors as a needle plot in Figure \ref{fig_signalrecovery} for the setting of $(M,N,G,k,\sigma_0)=(30,50,10,10,1)$. In this simulation we can already see that the Bayesian approaches (\texttt{dogss}, \texttt{ssep}, \texttt{bsgsss}) do a good job of reconstructing the correct signals. The group lasso \texttt{gglasso} chooses whole groups without within-group sparsity and as such has many false positives. The standard \texttt{lasso} and the sparse-group lasso \texttt{sgl} choose mostly correct coefficients, but underestimate their values.

\subsection{Influence of \texorpdfstring{$M, N, G$}{M, N, G} and \texorpdfstring{$k$}{k}}
First we analyze the performance on three conceptually different sets of parameters $(M,N,G,k)$ with fixed $\sigma_0=1$ and columns of $X$ drawn independently.

We evaluate the results of the methods on the simulated data on these aspects:
\begin{itemize}
\item boxplots of the AUROC and AUPR from 100 simulations with the same set of parameters,
\item boxplots of the relative prediction error on 100 additionally generated observations (test sample).
\end{itemize}

The choice of parameters $(M,N,G,k,\sigma_0)$ for the different settings of this section and the following ones is given in Table \ref{tab_sim1_parameters}.

\begin{table}
\centering
\begin{tabular}{ r | c c c c c c}
   & {$M$} & {$N$} & {$G$} & {$k$} & {$\sigma_0$} & corr. structure \\
  \hline
  \hline
  small & 30 & 30 & 5 & 5 & 1 & independent \\
  \hline
  medium & 30 & 100 & 20 & 10 & 1 & independent  \\
  \hline
  large & 100 & 1000 & 100 & 10 & 1 & independent \\
  \hline
  \multirow{5}{*}{noise} & \multirow{5}{*}{30} & \multirow{5}{*}{100} & \multirow{5}{*}{20} & \multirow{5}{*}{10} & 0 & \multirow{5}{*}{independent} \\
  & & & & & 0.1 & \\
  & & & & & 1 & \\
  & & & & & 3 & \\
  & & & & & 5 & \\
  \hline
  \multirow{3}{*}{correlation} & \multirow{3}{*}{30} & \multirow{3}{*}{100} & \multirow{3}{*}{20} & \multirow{3}{*}{10} & \multirow{3}{*}{1} & independent \\
  & & & & & & pairwise \\
  & & & & & & groupwise \\
\end{tabular}
\caption[Simulation settings for signal recovery.]{Choice of parameters for the different simulation settings in signal recovery.}\label{tab_sim1_parameters}
\end{table}

\begin{figure}
\centering
\includegraphics[width=\textwidth]{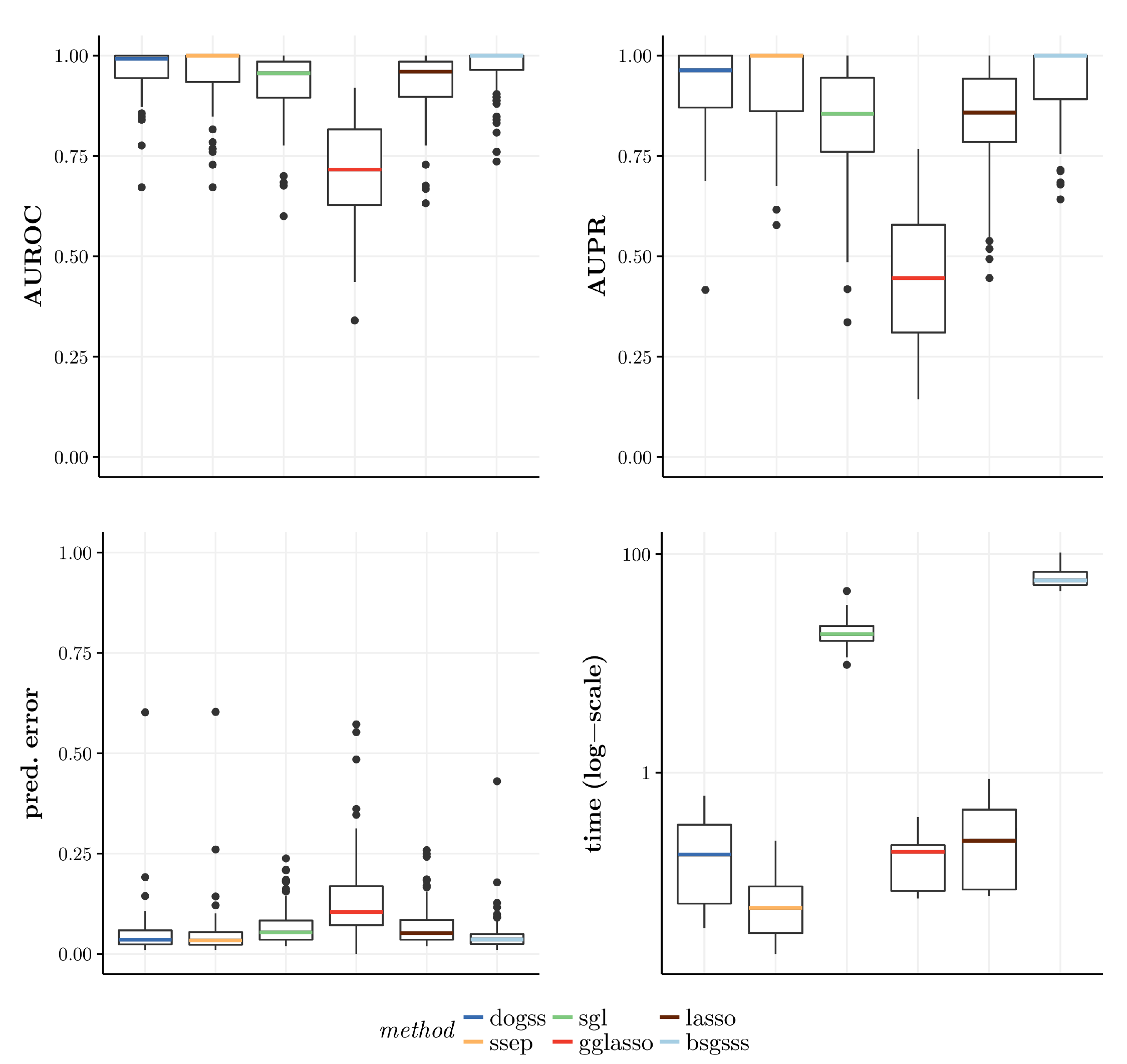}
\caption[Boxplots of AUROC, AUPR, prediction error and computing time for scenario 1.]{Boxplots of AUROC, AUPR, prediction error and computing time for scenario 1: results for $(M,N,G,k,\sigma_0)=(30,30,5,5,1)$ over 100 runs.}\label{fig_nocorr1}
\end{figure}

The simulation setting ``small'' of $(M,N,G,k,\sigma_0)=(30,30,5,5,1)$ describes a scenario where we have enough observations at hand ($M=N$) for predicting all of the coefficients, but we still have sparsity on the between-group and within-group level ($G=5$, $k=5$). Figure \ref{fig_nocorr1} shows the results, aggregated from 100 simulations. The AUROC for all methods except the group lasso is high (almost 1 in many cases). The Bayesian methods perform slightly better than the standard lasso and sparse-group lasso. In regard of the AUPR measure, this trend is more remarkable, while the group lasso performs worse. The group lasso does not take within-group sparsity into account and as such chooses too many coefficients which end up as false positives, we will see this behavior in all subsequent simulation scenarios. On the prediction error measure, we can see the same trends, with the Bayesian methods showing best results, while sparse-group lasso and lasso are close and group lasso performing worst. In this ``small'' simulation setting there is almost no difference if we include the grouping information or not: comparison of \texttt{dogss} versus \texttt{ssep} and \texttt{sgl} versus \texttt{lasso} shows basically no difference. We see important differences in the computing time: the Gibbs sampling approach \texttt{bsgsss} takes roughly 1000 times as long as the expectation propagation based calculations and the group lasso or standard lasso. The sparse-group lasso is approximately 100 times slower than the standard lasso.

The cutoff values for the prediction for all methods except \texttt{bsgsss} are derived by 10-fold cross validation with the 1se-rule. The predicting coefficients for the \texttt{bsgsss} method are recovered as the median values of the coefficients from the MCMC simulations.

\begin{figure}
\centering
\includegraphics[width=\textwidth]{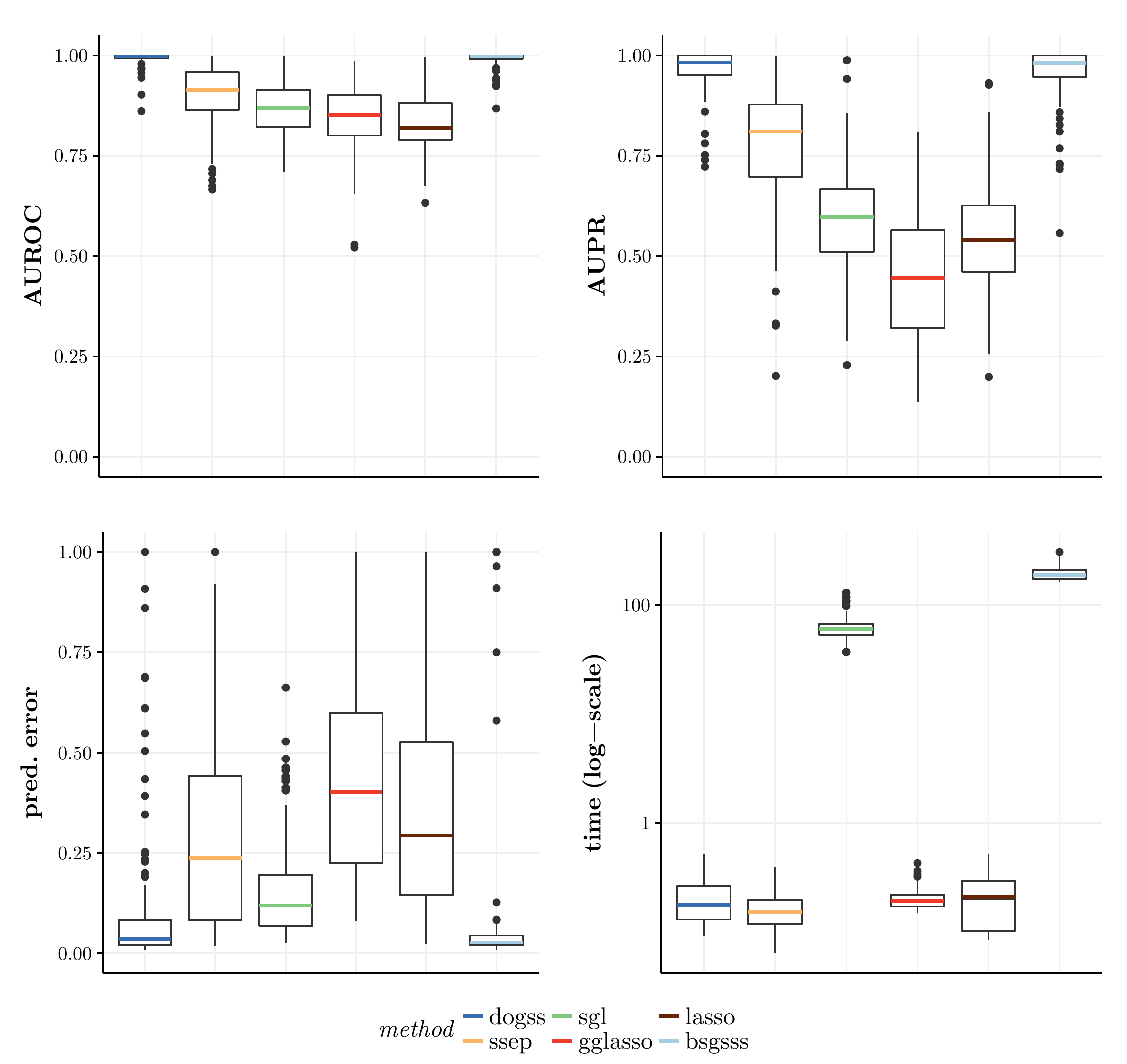}
\caption[Boxplots of AUROC, AUPR, prediction error and computing time for scenario 2.]{Boxplots of AUROC, AUPR, prediction error and computing time for scenario 2: results for $(M,N,G,k,\sigma_0)=(30,100,20,10,1)$ over 100 runs.}\label{fig_nocorr2}
\end{figure}

The simulation setting ``medium'' of $(M,N,G,k,\sigma_0)=(30,100,20,10,1)$ describes a scenario where we have more features than observations. We can see some important differences between the methods in regards of the AUROC, AUPR and prediction error measures: the two Bayesian approaches with two levels of group sparsity (\texttt{dogss} and \texttt{bsgsss}) perform the best and nearly identical. The lasso methods perform worse on the AUROC and AUPR measures than the Bayesian approaches, but the sparse-group lasso actually has a low prediction error (while not as low as the prediction errors of \texttt{dogss} and \texttt{bsgsss}). Again, the two expectation propagation based methods (\texttt{dogss} and \texttt{ssep}) as well the group lasso and standard lasso have low run times, while the sparse-group lasso \texttt{sgl} and the Gibbs sampling based approach \texttt{bsgsss} take about three orders of magnitude longer. There is also a wide range of the prediction errors with outliers on the prediction error boxplots for all methods.

\begin{figure}
\centering
\includegraphics[width=\textwidth]{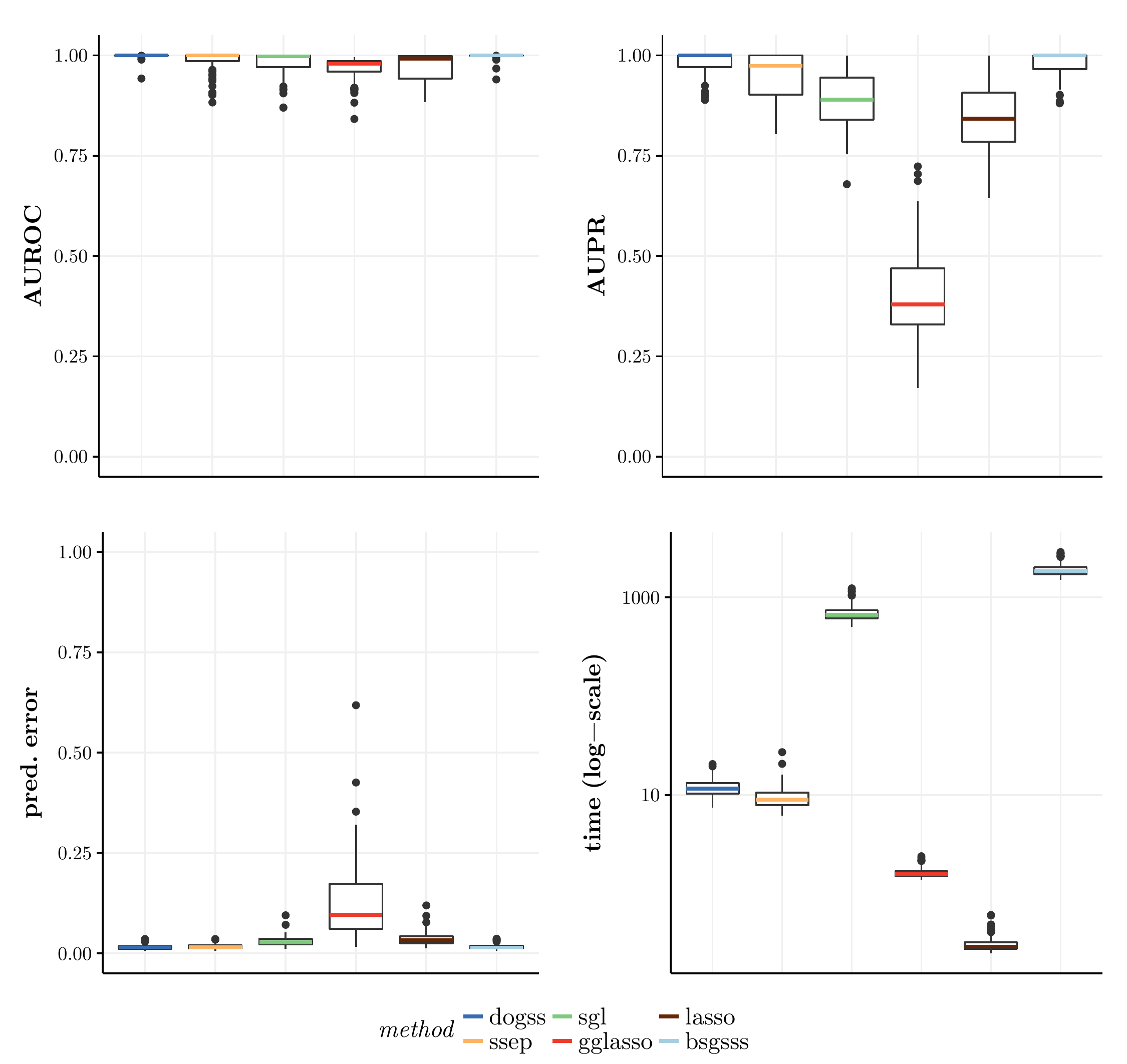}
\caption[Boxplots of AUROC, AUPR, prediction error and computing time for scenario 3.]{Boxplots of AUROC, AUPR, prediction error and computing time for scenario 3: results for $(M,N,G,k,\sigma_0)=(100,1000,100,10,1)$ over 100 runs.}\label{fig_nocorr3}
\end{figure}

The simulation setting ``large'' of $(M,N,G,k,\sigma_0)=(100,1000,100,10,1)$ describes a scenario where we have more features than observations and the number of features is high, while the signal is extremely sparse ($N=1000$, $k=10$). The trends from scenario ``medium'' regarding the AUROC and AUPR are carried forward: our approach \texttt{dogss} and the Gibbs sampling \texttt{bsgsss} perform best, with the standard spike-and-slab \texttt{ssep} in second place, followed by sparse-group lasso and standard lasso close up, whereas the group lasso performs worst. AUROC measures are high for all methods, but differences are more pronounced on the AUPR measure. The better results of our method \texttt{dogss} come at the price of increased run time compared to standard lasso (approximately two orders of magnitude), but it is still faster than the Gibbs sampling approach or the sparse-group lasso, which are in turn about two orders of magnitude slower than the expectation propagation based methods. The sparse-group method \texttt{dogss} performs slightly better than \texttt{ssep} (which does not take grouping information into account), but the effect is not as big as in the ``medium'' sized scenario.

\subsection{Influence of noise}\label{sec:noise}
\begin{figure}
\centering
\includegraphics[width=\textwidth]{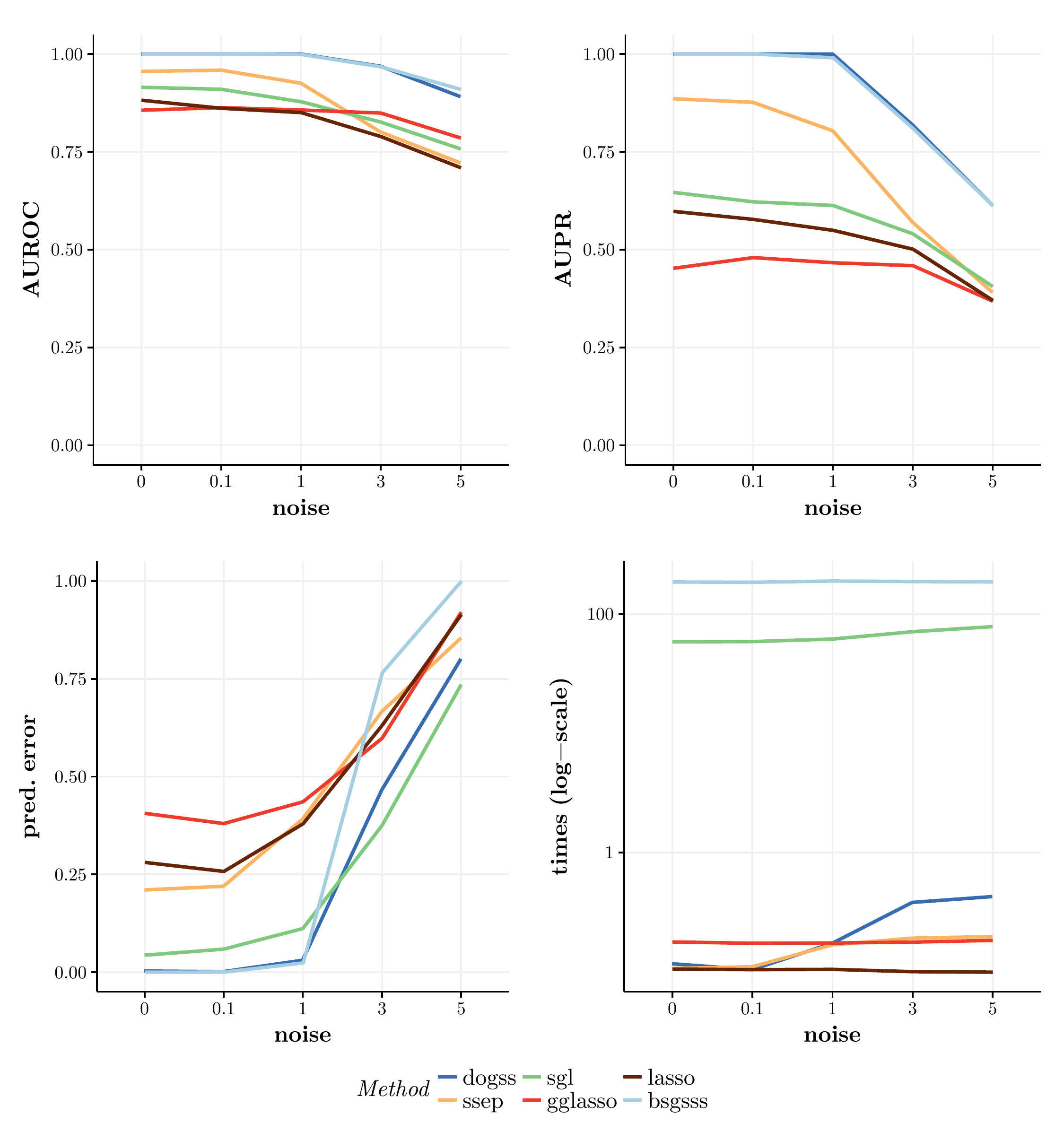}
\caption[AUROC, AUPR, prediction error and computing time for different levels of $\sigma_0$.]{AUROC, AUPR, prediction error and computing time for different levels of $\sigma_0$: median values over 100 runs for different methods with $(M,N,G,k)=(30,100,20,10)$.}\label{fig_noise}
\end{figure}

Second we evaluate the influence of $\sigma_0$ on a fixed set of parameters. To this end, we simulated 100 data sets like in the previous section with $(M,N,G,k)=(30,100,20,10)$ (the ``medium'' sized scenario), added noise with $\sigma_0 \in \lbrace  0; 0.1; 1; 3; 5 \rbrace$ and compared the performance of the six methods. The input parameter for $\sigma_0$ for our proposed method and the standard spike-and-slab was set to $1$ in all cases. For every data set, we measured AUROC, AUPR, prediction error and run time like before, we calculated the median values and show these in Figure \ref{fig_noise}.

We see that the correct specification of the noise parameter is important for our proposed algorithm and the standard spike-and-slab, too. If the provided noise parameter is higher than or equal to the actual one, our proposed algorithm gives good results. If the actual noise level is too high, the expectation propagation based methods suffer considerably (which can be seen most clearly on the AUPR measure). The lasso methods do not depend on the specification of the noise parameter, but their performance deteriorates for increasing noise levels, too, but not as steep as the Bayesian approaches. The grouped spike-and-slab implementation (\texttt{bsgsss}) with Gibbs sampling does not depend on the specification of the noise parameter either (since it samples this parameter from the data), but our simulations give a surprising result: in regards of the AUROC/AUPR measures, the Gibbs sampling performs similar to our proposed method, while the prediction error is actually worse for the \texttt{bsgsss} method for higher noise levels. The run time increases slightly for the expectation propagation based methods (\texttt{dogss} and \texttt{ssep}) with increasing levels of noise.

\subsection{Influence of correlated features}
\begin{figure}
\centering
\includegraphics[width=\textwidth]{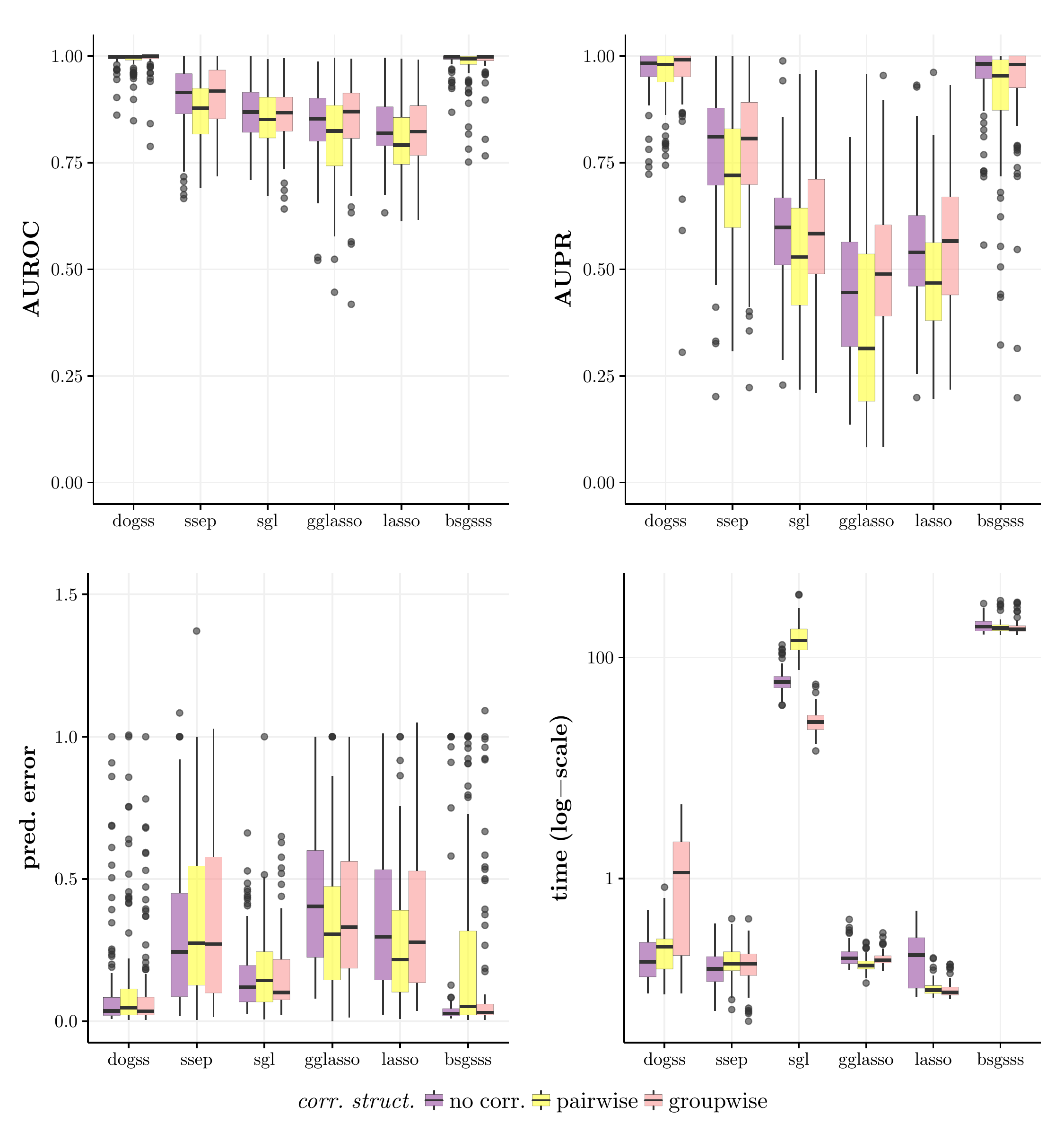}
\caption[Boxplots of AUROC, AUPR, prediction error and computing time  for different correlation structures.]{Boxplots of AUROC, AUPR, prediction error and computing time  for different correlation structures (uncorrelated, pairwise correlated or groupwise correlated features): results from 100 runs for different methods with $(M,N,G,k, \sigma_0)=(30,100,20,10, 1)$.}\label{fig_corrstruct}
\end{figure}

Third we assess the influence of the correlation structure between features within the data matrix $X$. The first structure we have already seen above, with columns of the data matrix $X$ drawn independently. The second structure ``pairwise'' refers to an overall pairwise correlation between features of $0.5$. The third structure ``groupwise'' is correlation on the group level: features within a group have a pairwise correlation of $0.5$, but every two features from different groups are independent. The parameters $(M,N,G,k, \sigma_0)=(30,100,20,10, 1)$ are chosen like in scenario ``medium''. The results of these simulations are aggregated in Figure \ref{fig_corrstruct}.

We cannot see many important differences between the performance in regard to the correlation structure. We see a small drop in AUROC/AUPR performance for all methods when features are pairwise correlated, while performance is roughly the same for independent or groupwise correlated features. Run times are different between correlation structures for the \texttt{dogss}, \texttt{sgl} and standard \texttt{lasso}.

\subsection{Influence of slab parameter}\label{sec:slabparameter}
Last, we evaluate the influence of $\sigma_{\textnormal{slab}}$ on a fixed set of $(M,N,G,k)$. Like the penalty parameter $\lambda$ for the lasso methods, the slab parameter for the spike-and-slab methods is a crucial value that needs to be chosen beforehand. But unlike the $\lambda$ parameter, which models the sparsity of the model directly and thus can and should be determined by cross-validation, the slab parameter rather puts a value to the expected size of the non-zero coefficients.

\begin{figure}
\centering
\includegraphics[width=0.8\textwidth]{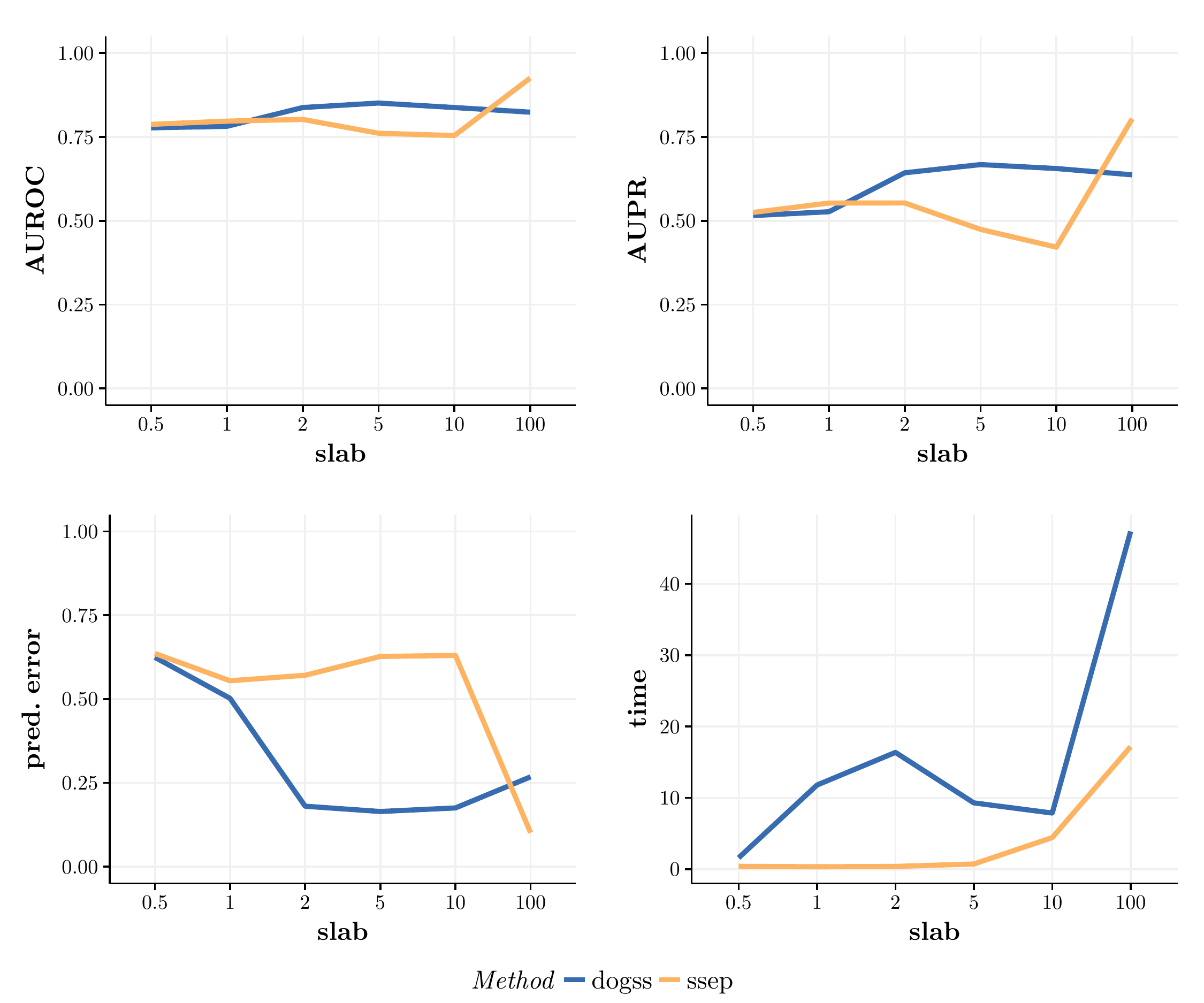}
\caption[Line plots of average AUROC, AUPR, prediction error and computing time for different $\sigma_{\textnormal{slab}}$ parameters.]{Line plots of average AUROC, AUPR, prediction error and computing time for different $\sigma_{\textnormal{slab}}$ parameters: results aggregated (median) over 100 runs for different methods with $(M,N,G,k, \sigma_0)=(30,100,20,10,1)$.}\label{fig_slab_analysis}
\end{figure}

We simulated 100 data sets like in the previous sections with \linebreak$(M,N,G,k,\sigma_0)=(30,100,20,10,1)$, that is the ``medium'' scenario, and compared the performance of our proposed approach and the standard spike-and-slab procedure. The input parameter for $\sigma_{\textnormal{slab}}$ for our proposed method and the standard spike-and-slab was chosen from $\lbrace 0.1;1;2;5;10;100 \rbrace$. For every data set, we measured AUROC, AUPR, prediction error and run time like before, we calculated the median values and show these in Figure \ref{fig_slab_analysis}.

The results for the AUROC and AUPR values are quite stable for our proposed method, with better results for higher values of the slab parameter. The prediction error is lowest for the values 2, 5 and 10. The run time increases a great deal for the value of 100. These results indicate that the slab parameter should be chosen reasonably in the range of the absolute size of the anticipated coefficients ($\beta$, here: the coefficients were drawn uniform-randomly from $\left[ -5;5 \right]$). We note that in this simulation scenario the standard spike-and-slab benefits from a rather high slab parameter (at the cost of increased run time).

\section{Network Reconstruction}

Network reconstruction is the problem of identifying a network graph that describes the dependencies between variables/features given some experimental measurements of these features. The network graph is a graphical model \citep{lauritzen_graphical_1996} consisting of nodes and edges between nodes, where nodes represent features and edges are the dependencies between the features. This graphical model is also called a Markov graph \citep[Chapter 17]{hastie_elements_2009}: two nodes are not connected by an edge if they are conditionally independent given all other nodes. Here we consider an undirected network graph, that is, edges do not indicate a causal direction (which would be represented as an arrow), but rather a mutual interdependence. 

If we assume that the data (that is, the $M$ observations of $P$ random variables $\mathbf{X}=(X_n)_{n=1}^P$) is generated from a multivariate normal distribution with covariance matrix $\Sigma$, the corresponding graphical model is called a Gaussian graphical model \citep[Section 8.1.4]{bishop_pattern_2006}: $
\mathbf{X} \sim \mathcal{N}_{P}(\mathbf{\mu}, \Sigma)$. Given this data matrix $\mathbf{X}$, we can reconstruct the generating network structure with a method called neighborhood selection \citep{meinshausen_high-dimensional_2006}: we consider $P$ independent feature selection problems, one for every column (feature) of $\mathbf{X}$, which corresponds to one particular node in the network graph. We determine this node's neighborhood of connected nodes, that is, find among the remaining $P-1=N$ variables the neighbors such that the current node is conditionally independent of all other variables given its neighborhood. In the network graph there are edges between the chosen node and all its neighborhood nodes, and no edges between nodes that are not neighbors. This means we perform $P=N+1$ instances of the sparse feature selection method of choice, where every instance consists of removing one column $i$ of the data matrix and using it as a response (dependent) variable $y=X_i$ that needs to be explained by the remaining $N$ variables (features, $X_{-i}=(X_j)_{j\neq i}$) as a standard regression problem:
\begin{align*}
X_i \equiv y &= X_{-i}\beta_i + \varepsilon_i \nlabel{eq:networkregression}\\
\textnormal{or more general: } y &= X \beta + \varepsilon.
\end{align*}

$X$ (or $X_{-i}$) is an $(M\times N)$-matrix, $y$ (or $X_i$) a vector with $M$ entries, $\beta$ (or $\beta_i$) a vector of coefficients with $N$ entries and $\varepsilon$ (or $\varepsilon_i$) a vector of $M$ entries that describes the stochastic error. Note that $X$ describes here now the reduced data matrix of dimensions $(M\times N)$. Equation \eqref{eq:networkregression} is the matrix notation of the equivalent formulation as element-wise linear sums:
\begin{align*}
x_{mi} \equiv y_{m} = \sum_{j\neq i} \beta_{ij}\cdot x_{mj} + \varepsilon_{mi},\ m=1,\ldots,M.
\end{align*}
Put together, we cast the problem of reconstructing a network from a data matrix into the problem of running $P$ separate feature selections, where the chosen features correspond to the retrieved edges in the network. Additionally, prior information in the form of a grouping of the features can be taken account of, too. We apply our proposed sparse-group Bayesian feature selection with expectation propagation as well as the other methods that we compared to in Section 3.

\subsection{Simulated networks}

To study the ability of our algorithm to reconstruct gene regulatory networks, we generate random network graphs with a known structure as Gaussian graphical models, which we compare as a gold standard to the results of our algorithms. A network graph where edges are just drawn discreetly-uniform and independent from the set of all possible edges is not a good representation of a biological network. Given the number of nodes and the number of edges, this random graph is also known as the model of \citet{erdos_random_1959}. A better model for a gene regulatory network is a scale-free network, where some genes (transcription factors or hubs) are connected to many other genes, while most genes are connected to only few other genes. That is, the out-degree of a node in this network follows approximately a power law $\mathbb{P}(k)=k^{-\gamma}$, where $k$ is the number of edges going out of a node and $\gamma$ is some positive constant. In our simulations we also allow for a hub to consist of multiple genes, which is a good representation of the behavior of biologically similar transcription factors. Gene regulatory networks (and many other networks arising in different contexts) show evidence for this scale-free topology structure, see for example \citep{clauset_power-law_2009} and \citep{babu_structure_2004}.

We simulate our own (approximately scale-free) network graphs based on four parameters (the number of nodes $P$, the number of groups $G$, the number of hub nodes $H$ and a random (Erdős–Rényi-like) edge probability $q$) from the following procedure:
\begin{enumerate}
\item we draw $G$ values $(p^{gs}_1,\ldots,p^{gs}_G)$ from $\mathcal{U}(0,1)$, these will represent the different group sizes (normalized such that $1=\sum_g p^{gs}_g$),
\item we assign $H$ hub nodes discreetly-uniform to all groups $\lbrace1,\ldots,G \rbrace$,
\item for every node $n$:
\begin{enumerate}
\item If it is not a hub node itself: sample a hub node from a group drawn from a categorical distribution $\operatorname{Cat}(p^{gs}_1,\ldots,p^{gs}_G)$ on $\lbrace1,\ldots,G \rbrace$ and draw an edge between node $n$ and the hub node.
\item If there are other hub nodes in this group, we draw additional edges to these hub nodes with probability 0.5 each.
\item Finally we add random edges to any other hub node in the whole graph with (low) probability $q$.
\end{enumerate}  
\end{enumerate}
The larger $q$, the more similar to a random Erdős–Rényi model the network gets, while with $q=0$ groups of nodes around their respective hubs are perfectly separated from each other.

\begin{figure}
\centering
\includegraphics[width=0.7\textwidth]{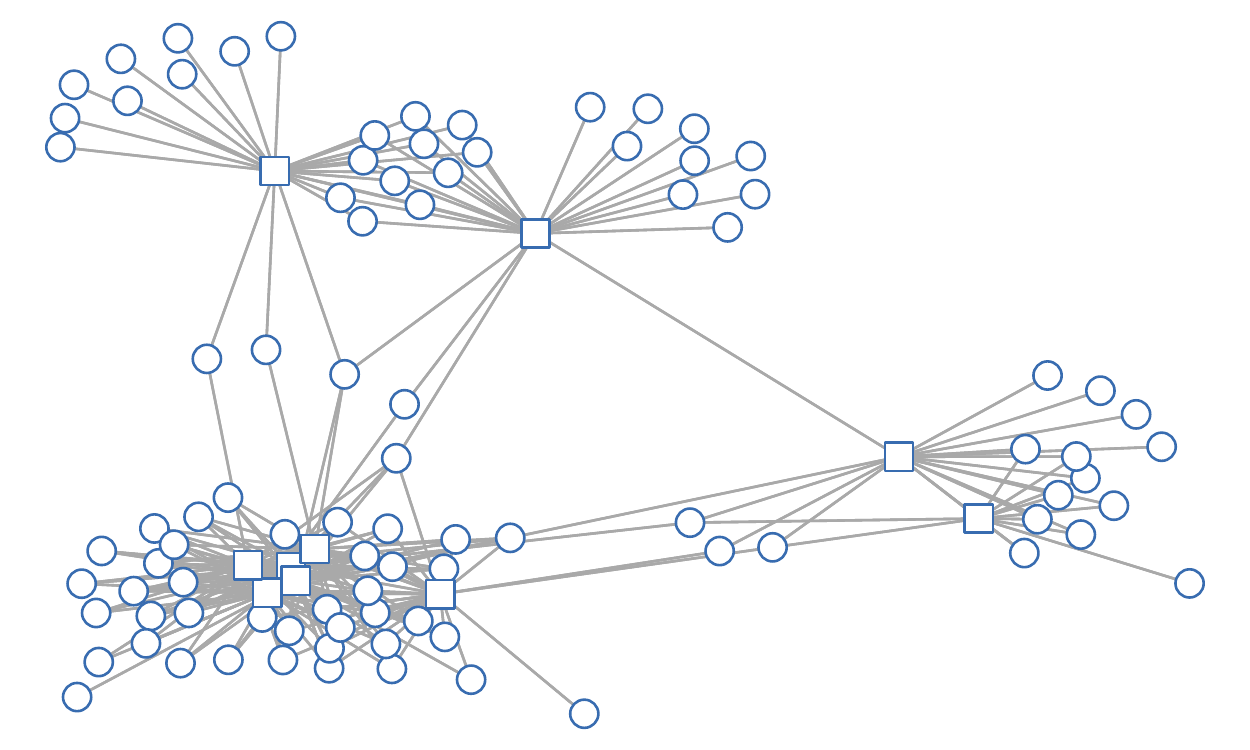}
\caption[Exemplary network generated from simulation setup.]{An exemplary network generated from parameters $(P,G,H,q)=(100,3,10,0.01)$. Squares correspond to hub nodes/transcription factors.}\label{fig_examplesimnetwork}
\end{figure}

An example network graph for a randomly generated network with parameters $(P,G,H,q)=(100,3,10,0.01)$ can be seen in Figure \ref{fig_examplesimnetwork}.

\begin{table}
\centering
\begin{tabular}{ r | c c c c}
   & {$P$} & {$G$} & {$H$} & {$q$}\\
  \hline
  \hline
  small & 100 & 3 & 10 & 0.01 \\
  \hline
  large & 1000 & 20 & 100 & 0.001 \\
\end{tabular}
\caption[Simulation settings for network reconstruction.]{Choice of parameters for the different simulation settings in network reconstruction.}\label{tab_sim2_parameters}
\end{table}

We ran simulations on two different network scenarios which differ in size, the parameters can be seen in Table \ref{tab_sim2_parameters}. For each scenario, we generated 100 random network graphs according to the procedure above. For each graph, we generated 100 random observations as a training set and an additional 100 observations as a test set with the \texttt{qpgraph} package from \citet{castelo_reverse_2009} in R, which also provides a precision matrix corresponding to the graph and observations. For each graph, we applied four different setups for the network reconstruction regarding the size of $X$ and the mapping $\mathcal{G}$: 
\begin{enumerate}
\item only hub nodes (transcription factors) are considered as features, with original grouping provided as in the graph generation,
\item all nodes (genes) are considered as features, with original grouping of the transcription factors as in the graph generation and additional groups for the non-hub nodes,
\item all nodes considered as features with a completely random grouping of the features.
\end{enumerate}
We compare the same methods like in the signal recovery simulations, with the exception of \texttt{bsgsss}: the Gibbs sampling approach takes too much computing time to be implemented in a feasible way for network reconstruction. We could include the \texttt{sgl} method by allowing its implementation to use its default relaxed values (threshold for convergence of $10^{-3}$, maximum number of iterations 1000 and a $\lambda$ sequence of just 20 values). The \texttt{gglasso} and \texttt{lasso} method use the same threshold and maximum number of iterations, but with the default length 100 of the $\lambda$ sequence. Our implementations \texttt{dogss} and \texttt{ssep} use the same threshold for convergence of $10^{-3}$, too, and a maximum of 100 iterations.

Additionally we test the predictive performance of the network reconstruction. To this end, we generate additional test data $X_{\textnormal{test}}=(x_{mp})$, $m=1,\ldots,100$, $p=1,\ldots,P$ and calculate the prediction error with the retrieved regression coefficient matrix $\hat{B}=(\hat{\beta}_{p_1p_2})$, $p_1,p_2=1,\ldots,P$ with $\hat{\beta}_{pp}=0$, from the neighborhood selection framework:
\begin{align*}
E = \frac{\sum_{m=1}^{100} \sum_{p_1=1}^{P} (x_{mp_1} - \sum_{\substack{p_2=1 \\ p_2\neq p_1}}^{P}\beta_{p_1p_2} x_{mp_2} )^2}{\sum_{m=1}^{100} \sum_{p=1}^{P}x_{mp}^2}.
\end{align*}

\begin{figure}
\centering
\includegraphics[width=\textwidth]{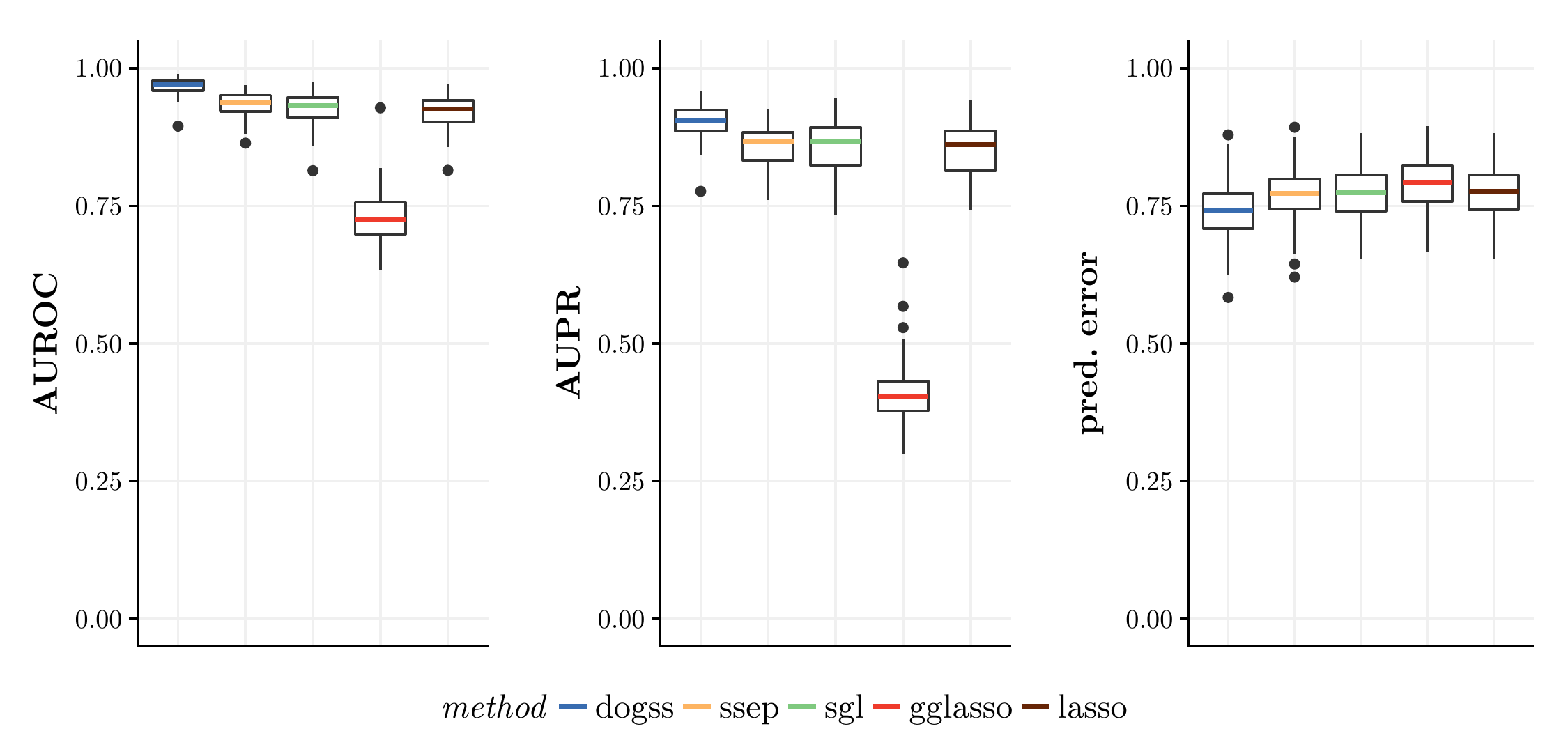}
\caption[Boxplots of AUROC/AUPR/prediction error on small simulated networks.]{Boxplots of AUROC/AUPR/prediction error for different methods running on 100 simulated small networks, only hub features considered.}\label{fig_network1}
\end{figure}

Figure \ref{fig_network1} shows the aggregated results for the small networks. Regarding the AUROC and AUPR measures, we see that the group lasso is clearly outperformed by all other methods. Furthermore, the differences between the methods without grouping information (\texttt{ssep} and \texttt{lasso}) are marginal, but also the \texttt{sgl} method which explicitly models sparsity on the between- and within- group level does not show any advantage over the \texttt{ssep} and the regular \texttt{lasso}. Our new method \texttt{dogss} performs best by a narrow margin. The same holds true for the prediction error measure on the held-out data.

\begin{figure}
\centering
\includegraphics[width=\textwidth]{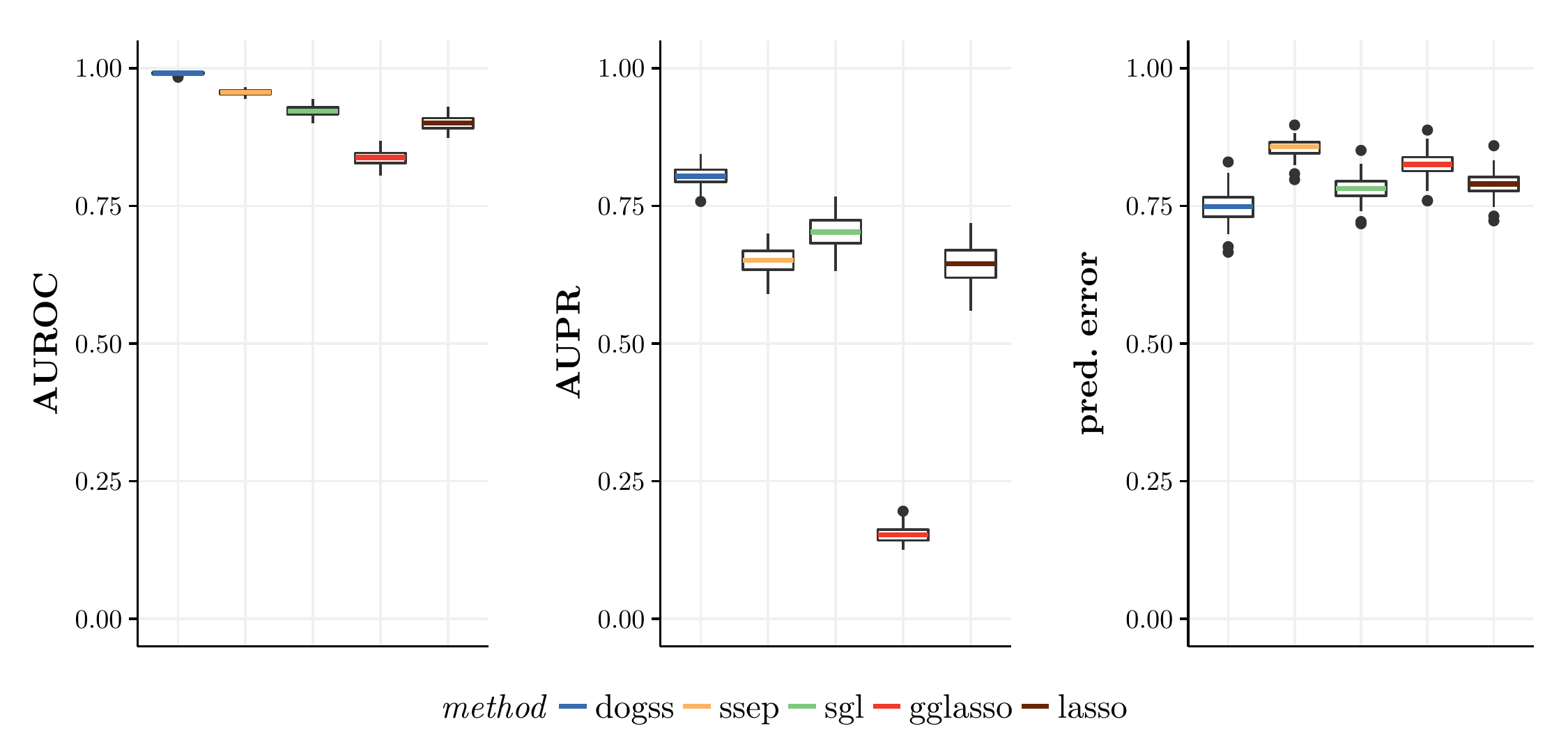}
\caption[Boxplots of AUROC/AUPR/prediction error on large simulated networks.]{Boxplots of AUROC/AUPR/prediction error for different methods running on 100 simulated large networks, only hub features considered.}\label{fig_network2}
\end{figure}

The results for the large networks are more diverse, see Figure \ref{fig_network2}. Again, the \texttt{gglasso} method suffers greatly on the AUROC and AUPR measures compared to all other methods, but it does not fare too bad on the prediction error measure. The comparison of the methods without grouping information ends in a tie: while \texttt{ssep} does better on the AUROC measure, \texttt{lasso} performs better with prediction error (and both have approximately equal AUPR). The sparse-group lasso \texttt{sgl} does a bit better than the standard \texttt{lasso}, but is clearly outperformed by our method \texttt{dogss} on all measures.

In the next step we compare the performance of the methods when we include all features for the neighborhood selection approach, and as such increase the sparsity. Additionally we compare the performance when features are grouped randomly opposed to the original grouping.

\begin{figure}
\centering
\includegraphics[width=\textwidth]{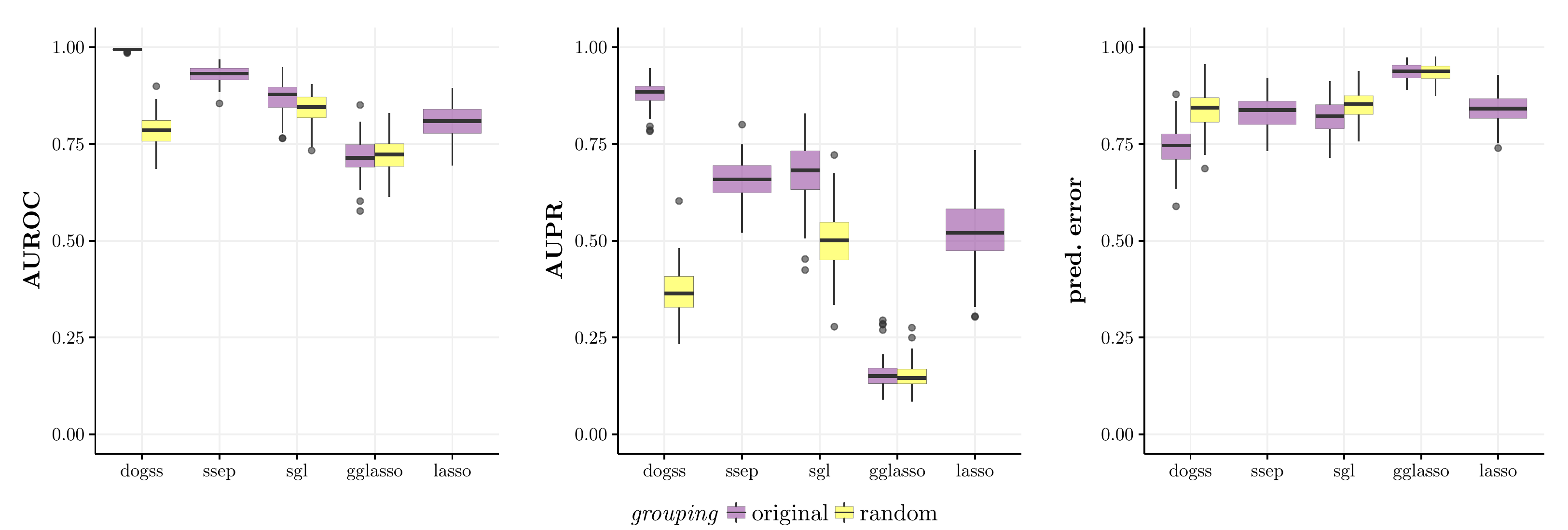}
\caption[Boxplots of AUROC/AUPR/prediction error on simulated small networks for different modes of grouping.]{Boxplots of AUROC/AUPR/prediction error for different methods running on 100 simulated networks (small), all features considered, with original grouping or random grouping.}\label{fig_networkgrouping_small}
\end{figure}

Let us first consider the small network reconstruction problem, see Figure \ref{fig_networkgrouping_small}. With the original grouping, our method \texttt{dogss} outperforms all other methods on all three measures of AUROC, AUPR and prediction error. The sparse-group lasso \texttt{sgl} outperforms the standard \texttt{lasso}, but the Bayesian method without grouping information \texttt{ssep} appears as a reasonable choice, too. Results for our method \texttt{dogss} deteriorate greatly for the random grouping, while the \texttt{ssep} method becomes the better choice with performance still better than \texttt{sgl}, which is in turn on par with the standard \texttt{lasso}.

\begin{figure}
\centering
\includegraphics[width=\textwidth]{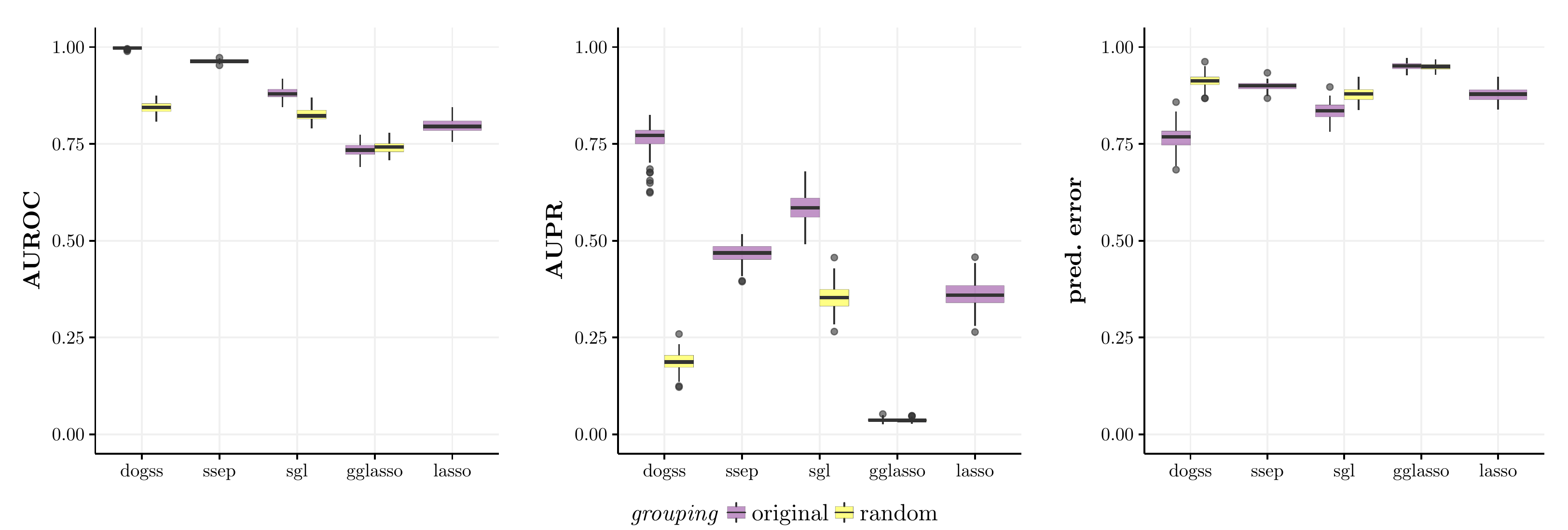}
\caption[Boxplots of AUROC/AUPR/prediction error on simulated large networks for different modes of grouping.]{Boxplots of AUROC/AUPR/prediction error for different methods running on 100 simulated networks (large), all features considered, with original grouping or random grouping.}\label{fig_networkgrouping_large}
\end{figure}

We see the same results from the small networks for the results on the large networks (Figure \ref{fig_networkgrouping_large}), but more pronounced. Our methods \texttt{dogss} performs best, but only if the grouping information is actually helpful. Otherwise, the Bayesian approach without grouping information \texttt{ssep} should be preferred.

\subsection{Results on biological data} \label{sec:biologicaldata}

One real-world example of a network reconstruction problem is the reconstruction of gene regulatory networks \citep{markowetz_inferring_2007}. A gene regulatory network consists of genes as nodes/features and edges as interactions between genes, where these edges are a simplistic shorthand for the complex molecular interplay of coding genetic sequences, RNA, and proteins \citep{karlebach_modelling_2008}.

Here we assess the algorithm's capability to reconstruct gene regulatory networks from real experimental data using prior information about the grouping of the variables. The DREAM5 gene network inference challenge \citep{marbach_wisdom_2012} provides extensive data along with a gold standard for the underlying networks. In the Escherichia coli challenge, 805 microarray experiments were given with measurements of gene expression for 4511 genes, of these 334 are transcription factors and as such candidate regulators for the genes. The gold standard network which was revealed after the challenge comprises of 2055 interactions within the network. To test the ability of our new algorithm (and the competing lasso approaches), we first derived a meaningful grouping of the transcription factors. This was not given to the participants of the challenge and as such our results are not directly comparable to the official results of the challenge.

We grouped the transcription factors by clustering them such that transcription factors belong to the same group if they predominantly bind to the same genes (thus, we identified the hubs), we will refer to this grouping as ``co-binding''. For comparison we also tested two additional groupings which do not contain any prior information and are as such purely data driven: grouping of features by simple kmeans clustering of gene expression and a random grouping of features. Finally we included a grouping derived from the supplementary files of \citet{marbach_wisdom_2012}, where a GO term analysis \citep{thegeneontologyconsortium_expansion_2017} gave groups of functional modules to all genes which we filtered for the transcriptions factors.

The total number of observations (805) is large and thus we decided to run the different methods on a (training) subset of randomly chosen 300 observations. This also leaves 505 observations as a test set for evaluation of the prediction error. We repeated the sampling of the training/test set 10 times and averaged the results.

\begin{figure}
\centering
\includegraphics[width=1\textwidth]{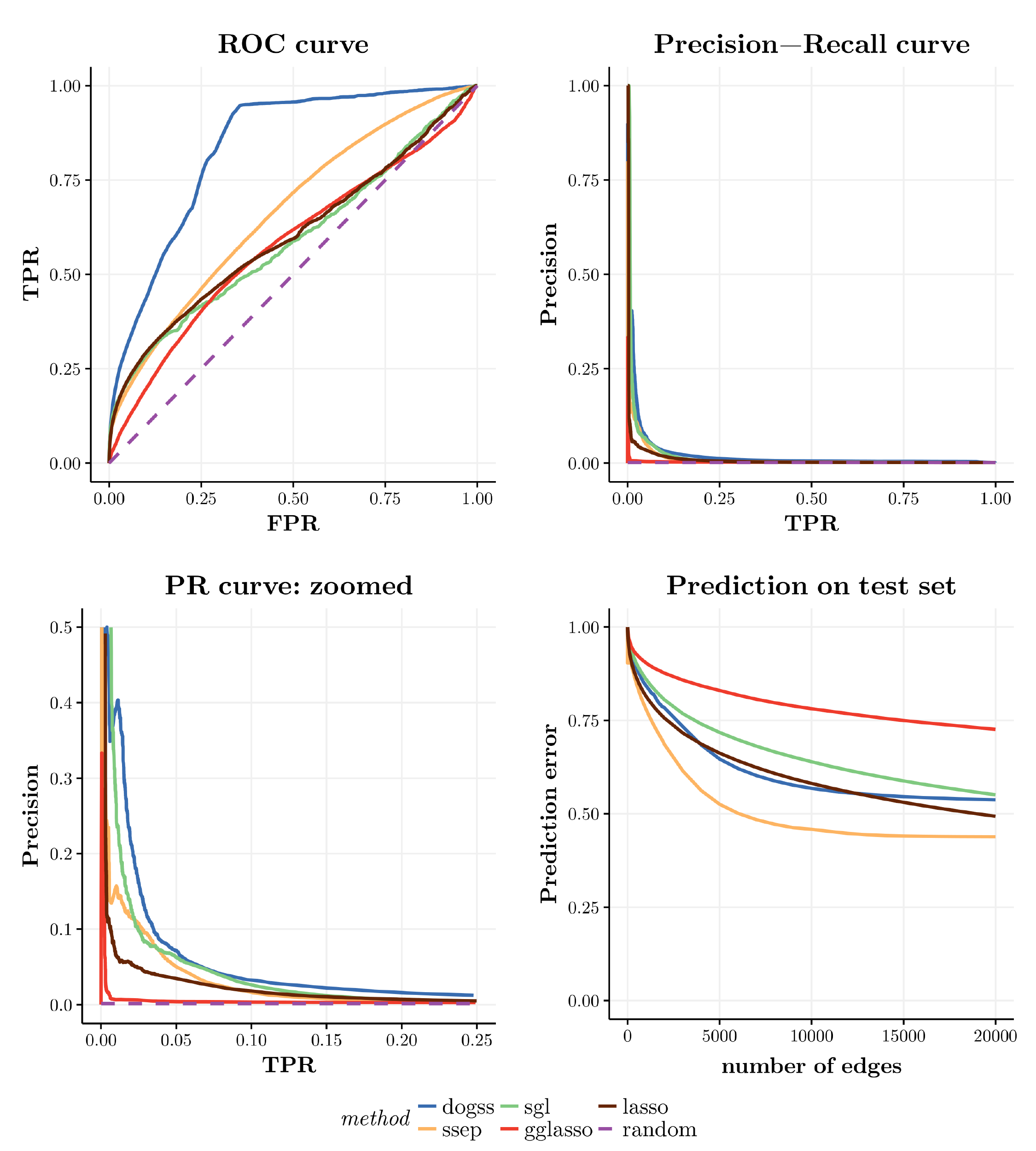}
\caption[ROC and Prec-Recall curves, prediction error on E. coli DREAM5.]{ROC and Prec-Recall curves, prediction error on E. coli DREAM5 data. The prediction error is given along the 20 000 top ranked edges for every method.}
\label{fig_ecoli}
\end{figure}

On each training set, we employed the neighborhood selection approach, that is, we ran every method for a total of $4511$ times (each gene once as a dependent response). We assess the quality of the network reconstruction based on the gold standard by deriving the true positive rate, false positive rate and precision along the ranked list of edges. The resulting lists were averaged over the ten runs and we plot the resulting average receiver operating characteristic (ROC) curve as well as the average precision-recall curve for the different algorithms (Figure \ref{fig_ecoli}). We also calculate the respective (average) area under the curve scores (Table \ref{tab_ecoli_area}).

\begin{table}
\centering
\begin{tabular}{ r | S S S S S S}
  method & {dogss} & {ssep} & {sgl} & {gglasso} & {lasso} & {random} \\
  \hline
  AUROC & 0.84 & 0.67 & 0.59 & 0.58 & 0.6 & 0.5 \\
  AUPR & 0.02 & 0.01 & 0.016 & 0.003 & 0.009 & 0.0014 \\
\end{tabular}
\caption[AUROC and AUPR for DREAM5 E. coli network reconstruction.]{AUROC and AUPR values for ROC and PR curves of the DREAM5 E. coli network reconstruction, see Figure \ref{fig_ecoli}.}\label{tab_ecoli_area}
\end{table}

The increase in the AUROC measure for our approach is immense, compared to alternative methods. That is, the prior information about the grouping of features is useful to identify more true positive edges without choosing too many false positive interactions. The second best method is the standard spike-and-slab, which does not use any grouping information. The results from the standard lasso and sparse-group lasso are barely distinguishable, indicating that the sparse-group lasso cannot take full advantage of the available grouping information. The group lasso is not a good candidate algorithm for this network reconstruction problem, its performance on the AUPR measure is quite poor.

The AUPR measure for all methods is far from close to the theoretical optimal value of 1, which is a common observation for ``real'' biological network reconstruction problems, see \citep{marbach_wisdom_2012}. All methods yield low AUPR values, indicating that the number of correctly identified interactions is in a bad proportion to the total number of called interactions. Our approach and the sparse-group lasso yield the highest AUPR values, and the standard spike-and-slab performs a little bit better than the standard lasso.

Furthermore we tested all algorithms on their predictive performance on this data set. We retrieved the regression coefficients from the neighborhood selection, chosen by cross-validation. For every method, we have a ranking of all possible edges along with a regression coefficient with every edge. This way we derive a prediction error curve along the ranked list: for rank $r$, we predict the expression of all 4177 genes that are not transcription factors by using the expression levels of the transcription factors in the held-out data. We sorted along the top $r$ edges respectively regression coefficients and set all other coefficients to 0. We measured the predictive performance via the relative sum of squared errors along the ranks $r$, that is
\begin{align*}
E(r) = \frac{\sum_{m=1}^{505} \sum_{n=1}^{4177} (y_{mn} - x_m \cdot \beta_{n}(r))^2}{\sum_{m=1}^{505} \sum_{n=1}^{4177}y_{mn}^2}.
\end{align*}
Figure \ref{fig_ecoli} shows the results for the different algorithms for the top 20 000 edges, averaged over the ten runs: we see that the standard spike-and-slab method makes the best prediction on the test data in the relevant realm of the top ranked edges (the number of true interactions is 2055). Our sparse-group Bayesian approach and the standard lasso perform roughly the same prediction-wise, while the sparse-group lasso eventually does better predictions than our approach, but only after approximately 20 000 included edges (the lines intersect outside of the figure margins). The group lasso is not useful for prediction in this setting.

\begin{table}
\centering
\begin{tabular}{ r | c c c c c c c c c }
  group & 1 & 2 & 3 & 4 & 5 & 6 & 7 & 8 & 9 \\
  \#TFs & 193 & 39 & 7 & 6 & 11 & 4 & 7 & 10 & 22 \\
  \hline
  \hline
    group & 10 & 11 & 12 & 13 & 14 & 15 & 16 & 17 & 18 \\
  \#TFs & 4 & 4 & 4 & 4 & 5 & 5 & 3 & 3 & 3 \\
\end{tabular}
\caption[Original grouping (co-binding) of DREAM5 E. coli transcription factors.]{Original grouping (co-binding) of DREAM5 E. coli transcription factors.}\label{tab_grouping}
\end{table}

\begin{figure}
\centering
\includegraphics[height=0.91\textheight]{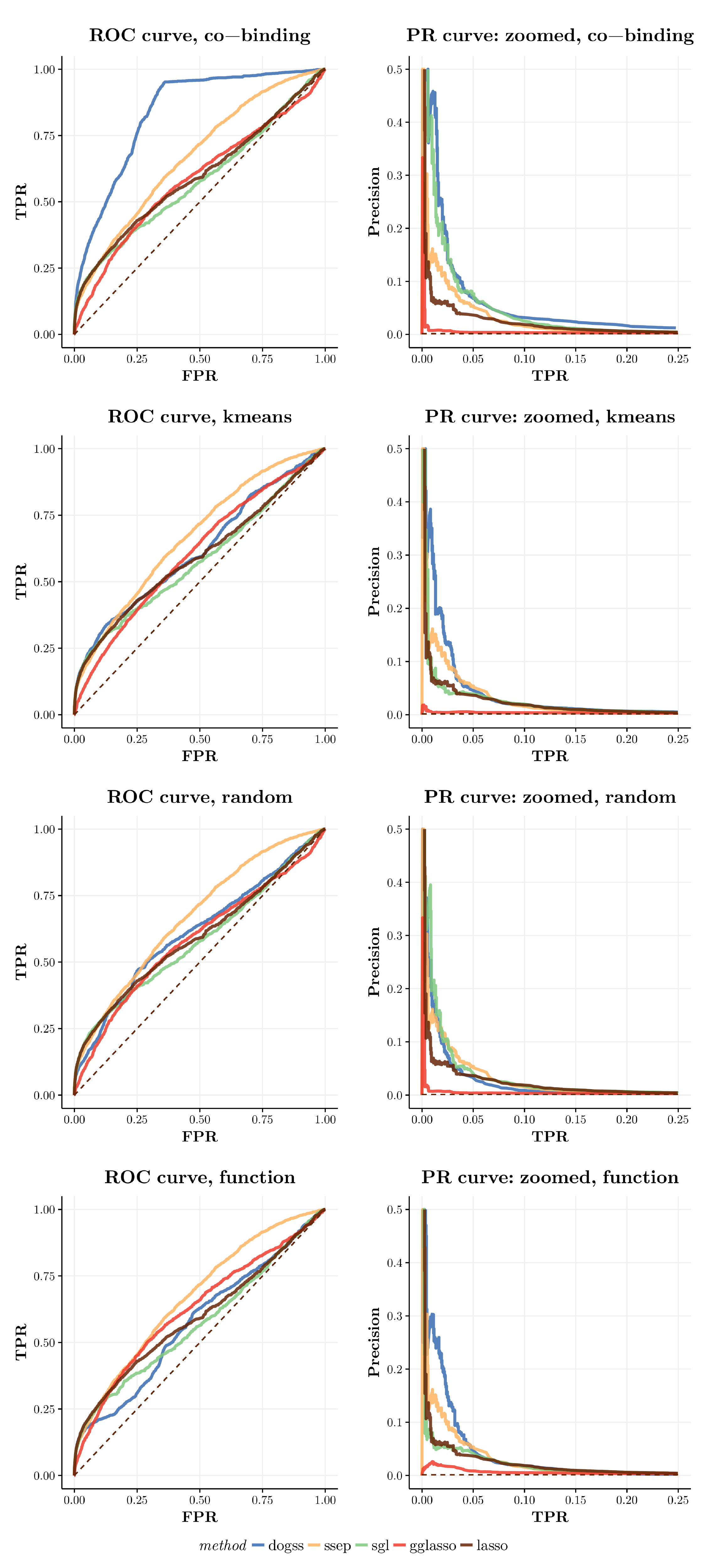} 
\caption[ROC and Prec-Recall curves for different groupings of features, DREAM5 data.]{ROC and Prec-Recall curves on E. coli DREAM5 data for different groupings: co-binding, clustering by kmeans, random, function.}
\label{fig_ecoli_diffgroups}
\end{figure}

Furthermore we compared our approach with the co-binding grouping of the transcription factors to a random grouping, a $k$-means based grouping, and a functional grouping \citep{marbach_wisdom_2012}. The grouping dubbed as co-binding was derived by grouping transcription factors that bind most often to the same genes, this led to 18 groups with different numbers of transcription factors (see Table \ref{tab_grouping} for the co-binding grouping). Now we assigned a random grouping to the transcription factors where group sizes are the same like in the co-binding grouping, that is we permuted the group memberships. Finally we grouped transcription factors by their gene expression similarity via $k$-means with $k=18$. We ran our algorithm again with these three new groupings and compare the results to the original grouping via the ROC and Precision-Recall curves, see Figure \ref{fig_ecoli_diffgroups}.

We observe that the co-binding grouping performs best in terms of ROC and Precision-Recall measure. Our sparse-group Bayesian method, the sparse-group lasso and the group lasso all perform worse than the standard spike-and-slab (with no grouping information) in regard of ROC curves for the kmeans, random and functional grouping. For the kmeans and functional grouping, our sparse-group Bayesian method performs best on the AUPR measure, while the sparse-group lasso performs worse than the standard spike-and-slab and approximately the same like the standard lasso. For the random grouping, sparse-group lasso, standard spike-and-slab and our method perform about the same.

\section{Discussion}
We developed a new feature selection algorithm based on a Bayesian framework that takes grouping information into account. We applied this method to the problems of signal recovery and (large-scale) network reconstruction from data.

The proposed Bayesian framework enforces sparsity on the between- and within-level for groups of features in tandem with a deterministic algorithm, expectation propagation, to derive the parameters of the framework. The Bayesian approach proved to be more reliable for feature selection with two-fold group sparsity than three different lasso methods (standard lasso, group lasso and sparse-group lasso). The expectation propagation algorithm as an alternative to the conventional Gibbs sampling is much faster while as accurate as Gibbs sampling, and as such feasible to apply to large network inference problems. To the best of our knowledge, this is the first application of Bayesian neighborhood selection with grouping information to the problem of network reconstruction.


Our comparisons indicate that a standard Bayesian approach without grouping information is a reliable choice under any circumstances. In the presence of helpful grouping information, our new approach of sparse-group Bayesian feature selection with expectation propagation outperforms all other methods. The standard lasso is the fastest method, but is outperformed by the Bayesian approaches and the sparse-group lasso. The group lasso suffers greatly in the presence of within-group sparsity. While the sparse-group lasso shows a slight advantage over the standard lasso in our analysis, this is by a narrow margin and not comparable to the gain of performance that our sparse-group Bayesian approach shows.

In the context of feature selection, lasso methods play a prominent role. We focused our work on Bayesian models based on the spike-and-slab, but interestingly, there are also Bayesian formulations of the lasso \citep{park_bayesian_2008}, group lasso \citep{kyung_penalized_2010}, and sparse-group lasso \citep{xu_bayesian_2015} respectively. All of these Bayesian lassos use Gibbs sampling to derive the parameters.


Extensive work was done on the theoretical foundations of feature selection with grouping information, but it was rarely applied to the problem of (gene) network reconstruction. Our work highlights the advantages of employing the sparse-group Bayesian framework on genetic data and network reconstruction. This was made possible only by inclusion of the expectation propagation algorithm, yielding faster run time compared to the much slower Gibbs sampling. While the expectation propagation algorithm was already applied to a standard model with no grouping \citep{hernandez-lobato_expectation_2015} and a between-group sparse model \citep{hernandez-lobato_generalized_2013}, it has not been extended to the two-level sparse group model yet. Our simulations and applications clearly show the superiority of this approach to detect correct sparsity patterns on a between- and within-group level.


\citet{li_bayesian_2017}, \citet{lin_joint_2017}, and \citet{atchade_scalable_2015} used Bayesian approaches for neighborhood selection, too, but all three works suffer from the use of Markov Chain Monte Carlo/Gibbs sampling algorithms, while not including grouping information either.




The standard lasso is a fast algorithm and does not depend on the specification of a noise parameter like our Bayesian approach. The problem of specifying the noise parameter in the Bayesian framework can be alleviated by cross-validation methods or plug-in estimators for the noise, although our simulations indicated that the specification of the noise gets problematic only in the realm of highly noisy data and the lasso methods suffer as much as the Bayesian methods in the presence of high noise.

Another advantage of the Bayesian approach over the lasso methods is that probabilities are provided along with the features, which makes it readily applicable to network reconstruction (since the probabilities give a natural and comparable ranking over different instances of the algorithm). \citet{meinshausen_stability_2010} and subsequently \citet{haury_tigress:_2012} presented a method called stability selection to assign probabilities to features within the lasso framework. Stability selection relies on repeating the lasso algorithm multiple times which increases the computational burden. We note that our algorithm can be used as a plug-in for the lasso methods within the same framework of stability selection, too, and further research should study if this is useful and how different feature selection methods perform with stability selection.

In general, grouping schemes and their impact on the performance of grouped feature selection methods merit further investigation. Another open question is how to determine the slab parameter beforehand. While this can be solved with cross-validation, (testable) slab parameter estimators need to be studied. Furthermore, ``fuzzy'' group memberships of features or features belonging to different groups at the same time \citep{zadeh_fuzzy_1965} are a natural property arising in biological data and it would be interesting to include this into the feature selection.


We note that our proposed method is not restricted to gene regulatory network inference or even on biological data for that matter. The framework is general and can be applied to any scenario where grouping information about the features is available. 



\appendix
\section*{Appendix A.}\label{app:theorem}
Here we derive the updates for all of the parameters of our spike-and-slab model within the expectation propagation framework.

Let $\mathcal{P}(\beta, Z, \Gamma)$ given by equations \eqref{eq:bayes}-\eqref{eq:bern_between} be the true posterior distribution and let $\mathcal{Q}(\beta, Z, \Gamma)$ given by equations \eqref{eq:myq}-\eqref{eq:qdistr} be an approximation to $\mathcal{P}$.

The expectation propagation algorithm matches expectations under $\mathcal{P}$ and $\mathcal{Q}$ of the sufficient statistics of $\mathcal{Q}$ (equation \ref{eq:EPmatchDogss}) by iteratively minimizing the Kullback-Leibler divergence $\operatorname{KL}(\mathcal{P}||\mathcal{Q})$. Since $\mathcal{P}$ and $\mathcal{Q}$ are factored (equations \ref{eq:myp} and \ref{eq:myq}), this is done by iterating through the factors $\tilde{f}_i$, $i=1, \ldots, 4$ and updating their respective parameters and $\mathcal{Q}$ in turns. In fact, only $i=2,3$ are considered, since $f_1$ and $\tilde{f}_1$ respectively $f_4$ and $\tilde{f}_4$ have the same form by choice. As such, the values for $\tilde{V}_1$ and $\tilde{m}_1$ (respectively $\tilde{V}_1^{-1}$ and $\tilde{V}_1^{-1}\tilde{m}_1$) as well as $\tilde{\varrho}_4$ do not need to be updated,  and thus their respective initial and final values are given by:
\begin{align*}
\tilde{V}_1^{-1} &= \frac{1}{\sigma_0^2}X^TX, \\
\tilde{V}_1^{-1}\tilde{m}_1 &= \frac{1}{\sigma_0^2}X^Ty,\\
\tilde{\varrho}_4 &= \varrho_0.
\end{align*}
This leaves us with the updates for $\tilde{f}_2$ and $\tilde{f}_3$. First, we cycle through the $\tilde{f}_{2,n}$ and find the parameters of the updated $\tilde{f}_{2,n}^{\textnormal{new}}$ via finding $\mathcal{Q}^{\backslash 2,n}$, then updating $\mathcal{Q}$ with the rules for the product of exponential family distributions. Second, the same is repeated for $\tilde{f}_{3}$ by cycling through the factors $\tilde{f}_{3,n}$ and finding the parameters of the updated $\tilde{f}_{3,n}^{\textnormal{new}}$ via finding $\mathcal{Q}^{\backslash 3,n}$ and afterwards updating $\mathcal{Q}$ like before.

Updates for $\tilde{f}_{2,n}$: The parameters $\tilde{V}_{n}^{\backslash 2,n}$, $\tilde{m}_{n}^{\backslash 2,n}$ and $\tilde{r}_n^{\backslash 2,n}$ of $\mathcal{Q}^{\backslash 2,n} \propto \mathcal{Q} \slash \tilde{f}_{2,n}$ are derived by using the rules for quotients of Bernoulli or normal distributions:
\begin{align*}
\tilde{V}_{n}^{\backslash 2,n} &= \left( \tilde{V}_{nn}^{-1} - \tilde{V}_{2,n}^{-1} \right)^{-1}, \\
\tilde{m}_{n}^{\backslash 2,n} &= \tilde{V}_{n}^{\backslash 2,n} \cdot \left( \tilde{V}_{nn}^{-1} \cdot \tilde{m}_n - \tilde{V}_{2,n}^{-1} \cdot \tilde{m}_{2,n} \right), \\
\tilde{r}_n^{\backslash 2,n} &= \tilde{r}_n - \tilde{r}_{2,n}.
\end{align*}
We find the updated $\tilde{f}_{2,n}^{\textnormal{new}}$ by minimizing the Kullback-Leibler divergence between $f_{2,n}\cdot \mathcal{Q}^{\backslash 2,n}$ and $\tilde{f}_{2,n}^{\textnormal{new}}\cdot \mathcal{Q}^{\backslash 2,n}$:
\begin{align*}
\mathrm{KL}(f_{2,n}\cdot \mathcal{Q}^{\backslash 2,n}\cdot \frac{1}{\mathsf{N}} || \tilde{f}_{2,n}^{\textnormal{new}} \cdot \mathcal{Q}^{\backslash 2,n} \cdot \frac{1}{\tilde{\mathsf{N}}}),
\end{align*}
where $\mathsf{N}$ and $\tilde{\mathsf{N}}$ are the appropriate normalizing constants.

Since we are only updating the marginal parameters $\tilde{r}_{2,n}^{\textnormal{new}}$, $\tilde{V}_{2,n}^{\textnormal{new}}$ and $\tilde{m}_{2,n}^{\textnormal{new}}$, we factorize $\mathcal{Q}^{\backslash 2,n}$ and $\mathsf{N}$ in respect to $n$ and will thus minimize
\begin{align*}
\mathrm{KL}(f_{2,n}\cdot \mathcal{Q}^{\backslash 2,n}_n &\cdot \frac{1}{\mathsf{N}_n} || \tilde{f}_{2,n}^{\textnormal{new}} \cdot \mathcal{Q}^{\backslash 2,n}_n \cdot \frac{1}{\tilde{\mathsf{N}}_n}) = \mathrm{KL}(\hat{\mathcal{P}}_n||\mathcal{Q}^{\textnormal{new}}_n), \nlabel{eq:klmargin2n} \\
\textnormal{where } f_{2,n}\cdot \mathcal{Q}^{\backslash 2,n}_n =& \left( Z_n \cdot \mathcal{N}(\beta_n|0,\sigma^2_{\textnormal{slab}})+(1-Z_n)\cdot \delta(\beta_n) \right) \cdot \mathcal{N}(\beta_n|\tilde{m}^{\backslash 2, n}_n, \tilde{V}^{\backslash 2, n}_n) \\
&\cdot \operatorname{Bern}(Z_n|\tilde{p}^{\backslash 2, n}_n) \cdot \operatorname{Bern}(\Gamma_{\mathcal{G}(n)}|\tilde{\pi}_{\mathcal{G}(n)}) \\
\textnormal{and thus } \mathsf{N}_n =& \int_{-\infty}^{+\infty} \sum_{Z_n=0,1} \sum_{\Gamma_{\mathcal{G}(n)}=0,1} f_{2,n}\cdot \mathcal{Q}^{\backslash 2,n}_n \mathrm{d}\beta \\
= & \tilde{p}^{\backslash 2, n}_n \cdot \mathcal{N}(\tilde{m}^{\backslash 2, n}_n, \tilde{V}^{\backslash 2, n}_n+\sigma^2_{\textnormal{slab}}) + (1-\tilde{p}^{\backslash 2, n}_n)\cdot \mathcal{N}(\tilde{m}^{\backslash 2, n}_n, \tilde{V}^{\backslash 2, n}_n).
\end{align*}
Minimizing \eqref{eq:klmargin2n} is the same as matching the expectations of the sufficient statistics $Z_n$, $\beta_n$ and $\beta_n^2$ under the probabilities from $\hat{\mathcal{P}}_n$ and $\mathcal{Q}^{\textnormal{new}}_n$, this gives us the updated parameters $\hat{p}^{\textnormal{new}}_n$, $\hat{m}^{\textnormal{new}}_n$ and $\hat{V}^{\textnormal{new}}_n$ of $\mathcal{Q}^{\textnormal{new}}_n$ (here we dropped all subscripts $_n$ or $_{2,n}$ and superscripts $^{\backslash 2,n}$ for better readability):
\begin{align*}
\hat{p}^{\textnormal{new}} \equiv \mathbb{E}_{\mathcal{Q}^{\textnormal{new}}}[Z] \stackrel{!}{=} \mathbb{E}_{\hat{\mathcal{P}}}[Z] &= \sum_{Z=0,1} \int_{-\infty}^{\infty} \sum_{\Gamma=0,1} Z\cdot \hat{\mathcal{P}}(\beta, Z, \Gamma) \mathrm{d}\beta \\
&= 0 + \frac{1}{\mathsf{N}}\mathcal{N}(\tilde{m}, \tilde{V}+\sigma^2_{\textnormal{slab}})\cdot \tilde{p} \cdot \tilde{\pi} + \frac{1}{\mathsf{N}}\mathcal{N}(\tilde{m}, \tilde{V}+\sigma^2_{\textnormal{slab}})\cdot \tilde{p} \cdot (1-\tilde{\pi}) \\
&= \frac{\tilde{p}}{\mathsf{N}} \cdot \mathcal{N}(\tilde{m}, \tilde{V}+\sigma^2_{\textnormal{slab}}),
\end{align*}
\begin{align*}
\hat{m}^{\textnormal{new}} \equiv \mathbb{E}_{\mathcal{Q}^{\textnormal{new}}}[\beta] \stackrel{!}{=} & \mathbb{E}_{\hat{\mathcal{P}}}[\beta] = \tilde{m} + \tilde{V} \cdot \frac{\partial}{\partial \tilde{m}} \mathsf{N}, \nlabel{eq:minkam} \\
\hat{V}^{\textnormal{new}} \equiv \mathbb{E}_{\mathcal{Q}^{\textnormal{new}}}[\beta^2] - \mathbb{E}_{\mathcal{Q}^{\textnormal{new}}}[\beta]^2 \stackrel{!}{=} & \mathbb{E}_{\hat{\mathcal{P}}}[\beta^2] - \mathbb{E}_{\hat{\mathcal{P}}}[\beta]^2 = \tilde{V} - \tilde{V}^2 \cdot \left( (\frac{\partial}{\partial \tilde{m}} \mathsf{N})^2 - 2\cdot \frac{\partial}{\partial \tilde{V}} \mathsf{N} \right).\nlabel{eq:minkaV}
\end{align*}
Equalities \eqref{eq:minkam} and \eqref{eq:minkaV} are from \citet[p. 15]{minka_family_2001}.

Remember, these are the new parameters of $\mathcal{Q}^{\textnormal{new}}_n$. To find the updated parameters of $\tilde{f}_{2,n}^{\textnormal{new}}$, we need to divide $\mathcal{Q}^{\textnormal{new}}_n$ by $\mathcal{Q}^{\backslash 2,n}_n$ and use the rules for the quotient of Bernoulli or normal distributions:
\begin{align*}
\tilde{r}_{2,n}^{\textnormal{new}} &= \hat{r}_{n}^{\textnormal{new}} - \tilde{r}_{n}^{\backslash 2, n}, \\
\tilde{V}_{2,n}^{\textnormal{new}} &= (1\slash \hat{V}_n^{\textnormal{new}} - 1\slash \tilde{V}_{n}^{\backslash 2, n})^{-1}, \\
\tilde{m}_{2,n}^{\textnormal{new}} &= \tilde{V}_{2,n}^{\textnormal{new}} \cdot (\hat{m}_n^{\textnormal{new}}\slash \hat{V}_n^{\textnormal{new}} - \tilde{m}_n^{\backslash 2, n}\slash \tilde{V}_{n}^{\backslash 2, n}).
\end{align*}
After some calculus and arithmetic (find the derivatives of $\mathsf{N}_n$ with respect to $\tilde{m}_n^{\backslash 2, n}$ and $\tilde{V}_{n}^{\backslash 2, n}$, then plug-in $\hat{p}^{\textnormal{new}}_n$, $\hat{m}^{\textnormal{new}}_n$ and $\hat{V}^{\textnormal{new}}_n$ and rearrange) we get the final analytical parameter updates of $\tilde{r}_{2,n}^{\textnormal{new}}$, $\tilde{V}_{2,n}^{\textnormal{new}}$ and $\tilde{m}_{2,n}^{\textnormal{new}}$:
\begin{align*}
\tilde{r}_{2,n}^{\textnormal{new}} &= \frac{1}{2} \cdot \left( \log \left( \frac{\tilde{V}_{n}^{\backslash 2,n}}{\tilde{V}_{n}^{\backslash 2,n}+\sigma^2_{\textnormal{slab}}} \right) + (\tilde{m}_{n}^{\backslash 2,n})^2 \cdot \left( 1\slash \tilde{V}_{n}^{\backslash 2,n} - 1\slash(\tilde{V}_{n}^{\backslash 2,n}+\sigma^2_{\textnormal{slab}}) \right) \right),\\
\tilde{V}_{2,n}^{\textnormal{new}} &= \frac{1}{a_n^2-b_n} - \tilde{V}_{n}^{\backslash 2,n},\\
\tilde{m}_{2,n}^{\textnormal{new}} &= \tilde{m}_{n}^{\backslash 2,n} - a_n\cdot (\tilde{V}_{2,n}^{\textnormal{new}} + \tilde{V}_{n}^{\backslash 2,n}),\\
\textnormal{with } a_n &= p^{\textnormal{aux}}_{n} \cdot \frac{\tilde{m}_{n}^{\backslash 2,n}}{\tilde{V}_{n}^{\backslash 2,n}+\sigma^2_{\textnormal{slab}}} + (1-p^{\textnormal{aux}}_{n}) \cdot \frac{\tilde{m}_{n}^{\backslash 2,n}}{\tilde{V}_{n}^{\backslash 2,n}}, \\
b_n &= p^{\textnormal{aux}}_{n} \cdot \frac{(\tilde{m}_{n}^{\backslash 2,n})^2-\tilde{V}_{n}^{\backslash 2,n}-\sigma^2_{\textnormal{slab}}}{(\tilde{V}_{n}^{\backslash 2,n}+\sigma^2_{\textnormal{slab}})^2} + (1-p^{\textnormal{aux}}_{n}) \cdot \frac{(\tilde{m}_{n}^{\backslash 2,n})^2-\tilde{V}_{n}^{\backslash 2,n}}{(\tilde{V}_{n}^{\backslash 2,n})^2}\\
\textnormal{and } p^{\textnormal{aux}}_{n} &= \operatorname{sigmoid}(\tilde{r}_{2,n}^{\textnormal{new}}+\tilde{r}_n^{\backslash 2,n}).
\end{align*}
These do not rely on the parameters of $\mathcal{Q}^{\textnormal{new}}_n$ anymore, so in practice one calculates the updates of $\tilde{f}_{2,n}^{\textnormal{new}}$ directly from the parameters of $\mathcal{Q}^{\backslash 2,n}$.

The updates for $\mathcal{Q}$ after updating $\tilde{f}_{2,n}^{\textnormal{new}}$ are derived from the rules for the product of Bernoulli and normal distributions:
\begin{align*}
\tilde{V} &= \left( \tilde{V}_1^{-1} + \left(\tilde{V}_2^{\textnormal{new}}\right)^{-1} \right)^{-1},\\
\tilde{m} &= \tilde{V} \left( \tilde{V}_1^{-1}\tilde{m}_1 + \left(\tilde{V}_2^{\textnormal{new}}\right)^{-1} \tilde{m}_{2}^{\textnormal{new}} \right), \\
\tilde{r}_n &= \tilde{r}_{2,n}^{\textnormal{new}} + \tilde{r}_{3,n},\\
\tilde{\varrho} &\textnormal{ does not change.}
\end{align*}
Updates for $\tilde{f}_{3,n}$: The parameters $\tilde{r}_n^{\backslash 3,n}$ and $\tilde{\varrho}_n^{\backslash 3,n}$ of $\mathcal{Q}^{\backslash 3,n} \propto \mathcal{Q} \slash \tilde{f}_{3,n}$ are derived by using the rules for quotients of Bernoulli distributions:
\begin{align*}
\tilde{\varrho}_{n}^{\backslash 3,n} &= \tilde{\varrho}_{\mathcal{G}(n)} - \tilde{\varrho}_{3,n},\\
\tilde{r}_{n}^{\backslash 3,n} &= \tilde{r}_n - \tilde{r}_{3,n}.
\end{align*}
We find the updated $\tilde{f}_{3,n}^{\textnormal{new}}$ by minimizing the Kullback-Leibler divergence between $f_{3,n}\cdot \mathcal{Q}^{\backslash 3,n}$ and $\tilde{f}_{3,n}^{\textnormal{new}}\cdot \mathcal{Q}^{\backslash 3,n}$:
\begin{align*}
\mathrm{KL}(f_{3,n}\cdot \mathcal{Q}^{\backslash 3,n}\cdot \frac{1}{\mathsf{N}} || \tilde{f}_{3,n}^{\textnormal{new}} \cdot \mathcal{Q}^{\backslash 3,n} \cdot \frac{1}{\tilde{\mathsf{N}}}),
\end{align*}
where $\mathsf{N}$ and $\tilde{\mathsf{N}}$ are the appropriate normalizing constants.

Since we are only updating the marginal parameters $\tilde{r}_{3,n}^{\textnormal{new}}$ and $\tilde{\varrho}_{3,n}^{\textnormal{new}}$, we factorize $\mathcal{Q}^{\backslash 3,n}$ and $\mathsf{N}$ with respect to $n$ and will thus minimize
\begin{align*}
\mathrm{KL}(f_{3,n}\cdot \mathcal{Q}^{\backslash 3,n}_n &\cdot \frac{1}{\mathsf{N}_n} || \tilde{f}_{3,n}^{\textnormal{new}} \cdot \mathcal{Q}^{\backslash 3,n}_n \cdot \frac{1}{\tilde{\mathsf{N}}_n}) = \mathrm{KL}(\hat{\mathcal{P}}_n||\mathcal{Q}^{\textnormal{new}}_n), \nlabel{eq:klmargin3n} \\
\textnormal{where } f_{3,n}\cdot \mathcal{Q}^{\backslash 3,n}_n =& \left( \Gamma_{\mathcal{G}(n)} \cdot \operatorname{Bern}(Z_n|p_{0,n})+(1-\Gamma_{\mathcal{G}(n)})\cdot \delta(Z_n) \right) \cdot \mathcal{N}(\beta_n|\tilde{m}^{\backslash 3, n}_n, \tilde{V}^{\backslash 3, n}_n) \\
&\cdot \operatorname{Bern}(Z_n|\tilde{p}^{\backslash 3, n}_n) \cdot \operatorname{Bern}(\Gamma_{\mathcal{G}(n)}|\tilde{\pi}^{\backslash 3, n}_n) \\
\textnormal{and thus } \mathsf{N}_n =& \int_{-\infty}^{+\infty} \sum_{Z_n=0,1} \sum_{\Gamma_{\mathcal{G}(n)}=0,1} f_{3,n}\cdot \mathcal{Q}^{\backslash 3,n}_n \mathrm{d}\beta \\
= & \tilde{\pi}^{\backslash 3, n}_n \cdot \left( \tilde{p}^{\backslash 3, n}_np_{0,n} + (1-\tilde{p}^{\backslash 3, n}_n)(1-p_{0,n}) \right) + (1-\tilde{\pi}^{\backslash 3, n}_n)\cdot (1-\tilde{p}^{\backslash 3, n}_n).
\end{align*}
Minimizing \eqref{eq:klmargin3n} is the same as matching the expectations of the sufficient statistics $Z_n$, $\beta_n$ and $\beta_n^2$ under the probabilities from $\hat{\mathcal{P}}_n$ and $\mathcal{Q}^{\textnormal{new}}_n$, this gives us the updated parameters $\hat{p}^{\textnormal{new}}_n$ and $\hat{\pi}^{\textnormal{new}}_n$ of $\mathcal{Q}^{\textnormal{new}}_n$ (here we drop all subscripts $_n$ or $_{3,n}$ and superscripts $^{\backslash 3,n}$ for better readability):
\begin{align*}
\hat{p}^{\textnormal{new}} \equiv \mathbb{E}_{\mathcal{Q}^{\textnormal{new}}}[Z] \stackrel{!}{=} \mathbb{E}_{\hat{\mathcal{P}}}[Z] &= \sum_{Z=0,1} \int_{-\infty}^{\infty} \sum_{\Gamma=0,1} Z\cdot \hat{\mathcal{P}}(\beta, Z, \Gamma) \mathrm{d}\beta \\
&= \frac{\tilde{p}\tilde{\pi}p_0}{\mathsf{N}} \\
\hat{\pi}^{\textnormal{new}} \equiv \mathbb{E}_{\mathcal{Q}^{\textnormal{new}}}[\Gamma] \stackrel{!}{=} \mathbb{E}_{\hat{\mathcal{P}}}[\Gamma] &= \sum_{Z=0,1} \int_{-\infty}^{\infty} \sum_{\Gamma=0,1} Z\cdot \hat{\mathcal{P}}(\beta, Z, \Gamma) \mathrm{d}\beta \\
&= \frac{\tilde{p}\tilde{\pi}p_0 +(1-\tilde{p})(1-p_0)\tilde{\pi}}{\mathsf{N}}.
\end{align*}
These are the new parameters of $\mathcal{Q}^{\textnormal{new}}_n$. To find the updated parameters of $\tilde{f}_{3,n}^{\textnormal{new}}$, we need to divide $\mathcal{Q}^{\textnormal{new}}_n$ by $\mathcal{Q}^{\backslash 3,n}_n$ and use the rules for the quotient of Bernoulli distributions:
\begin{align*}
\tilde{r}_{3,n}^{\textnormal{new}} &= \hat{r}_{n}^{\textnormal{new}} - \tilde{r}_{n}^{\backslash 3, n}, \\
\tilde{\varrho}_{3,n}^{\textnormal{new}} &= \hat{\varrho}_{n}^{\textnormal{new}} - \tilde{\varrho}_{n}^{\backslash 3, n}.
\end{align*}
After some arithmetic operations we get the final analytical parameter updates of $\tilde{\varrho}_{3,n}^{\textnormal{new}}$ and $\tilde{r}_{3,n}^{\textnormal{new}}$: 
\begin{align*}
\tilde{\varrho}_{3,n}^{\textnormal{new}} &= -\log(1-\tilde{p}_{n}^{\backslash 3,n}) + \log(\tilde{p}_{n}^{\backslash 3,n}\cdot p_{0,n} + (1-\tilde{p}_{n}^{\backslash 3,n}) \cdot (1-p_{0,n}))\\
&= \log(1+p_{0,n}\cdot(\exp(\tilde{r}_{n}^{\backslash 3,n})-1)),\\
\tilde{r}_{3,n}^{\textnormal{new}} &= \operatorname{logit}(\tilde{\pi}_{n}^{\backslash 3,n} \cdot p_{0,n}) \\
&= \log p_{0,n} - \log(1-p_{0,n}+\exp(-\tilde{\varrho}_{n}^{\backslash 3,n})).
\end{align*}
These do not rely on the parameters of $\mathcal{Q}^{\textnormal{new}}_n$, so in practice one calculates the updates of $\tilde{f}_{3,n}^{\textnormal{new}}$ directly.

The updates for $\mathcal{Q}$ after updating $\tilde{f}_{3,n}$ are derived from the rules for the product of Bernoulli distributions:
\begin{align*}
\tilde{V} &\textnormal{ does not change,} \\
\tilde{m} &\textnormal{ does not change,} \\
\tilde{\varrho}_{\mathcal{G}(n)} &= \tilde{\varrho}_{4,\mathcal{G}(n)} + \sum_{l:\mathcal{G}(l)=\mathcal{G}(n)} \tilde{\varrho}_{3,l}^{\textnormal{new}}, \\
\tilde{r}_n &= \tilde{r}_{2,n} + \tilde{r}_{3,n}^{\textnormal{new}}.
\end{align*}

\section*{Appendix B.}\label{app:rules_bernoulli_normal}

Since the Bernoulli distribution belongs to the exponential family of distributions, the product of two Bernoulli distributions is again a Bernoulli distribution up to a normalization constant, and the parameter of the new Bernoulli distribution can be calculated as follows, see also \citet[Appendix A.1]{hernandez-lobato_prediction_2009}:
\begin{align*}
\operatorname{Bern}(x|p_1) \cdot \operatorname{Bern}(x|p_2) \propto \operatorname{Bern}(x|p) \\
\Rightarrow p = \frac{p_1p_2}{p_1p_2+(1-p_1)(1-p_2)}.
\end{align*}
But with $r = \log \frac{p}{1-p} = \operatorname{logit}(p)$ we have:
\begin{align*}
r = r_1 + r_2,
\end{align*}
which is helpful from a numerical point of view.

Similar results hold true for the quotient of Bernoulli distributions:
\begin{align*}
\operatorname{Bern}(x|p_1) &\slash \operatorname{Bern}(x|p_2) \propto \operatorname{Bern}(x|p) \\
\Rightarrow p &= \frac{p_1 \slash p_2}{p_1 \slash p_2+(1-p_1)\slash(1-p_2)} \\
\Rightarrow r &= r_1 - r_2.
\end{align*}

The product or quotient of two normal distribution functions is again a normal distribution function, up to a normalization constant. The parameters $\mu$ and $\Sigma$ can be calculated from the parameters of the factor/denominator/numerator distributions as follows:
\begin{align*}
\mathcal{N}(x|\mu_1, \Sigma_1) \cdot \mathcal{N}(x|\mu_2, \Sigma_2) \propto \mathcal{N}(x|\mu, \Sigma) \\
\Rightarrow \Sigma = (\Sigma_1^{-1} + \Sigma_2^{-1})^{-1}, \\
\mu = \Sigma (\Sigma_1^{-1}\mu_1 + \Sigma_2^{-1}\mu_2).
\end{align*}
\begin{align*}
\mathcal{N}(x|\mu_1, \Sigma_1) \slash \mathcal{N}(x|\mu_2, \Sigma_2) \propto \mathcal{N}(x|\mu, \Sigma) \\
\Rightarrow \Sigma = (\Sigma_1^{-1} - \Sigma_2^{-1})^{-1}, \\
\mu = \Sigma (\Sigma_1^{-1}\mu_1 - \Sigma_2^{-1}\mu_2).
\end{align*}
The results for a univariate normal distribution are just the same with $\Sigma=\sigma^2$.

\vskip 0.2in
\bibliographystyle{abbrvnat}
\bibliography{bibliography}

\begin{thebibliography}{44}
\providecommand{\natexlab}[1]{#1}
\providecommand{\url}[1]{\texttt{#1}}
\expandafter\ifx\csname urlstyle\endcsname\relax
  \providecommand{\doi}[1]{doi: #1}\else
  \providecommand{\doi}{doi: \begingroup \urlstyle{rm}\Url}\fi

\bibitem[Atchadé(2015)]{atchade_scalable_2015}
Y.~Atchadé.
\newblock A scalable quasi-{Bayesian} framework for {Gaussian} graphical
  models.
\newblock \emph{arXiv preprint}, Dec. 2015.
\newblock arXiv: 1512.07934.

\bibitem[Babu et~al.(2004)Babu, Luscombe, Aravind, Gerstein, and
  Teichmann]{babu_structure_2004}
M.~M. Babu, N.~M. Luscombe, L.~Aravind, M.~Gerstein, and S.~A. Teichmann.
\newblock Structure and evolution of transcriptional regulatory networks.
\newblock \emph{Current Opinion in Structural Biology}, 14\penalty0
  (3):\penalty0 283--291, June 2004.
\newblock ISSN 0959-440X.
\newblock \doi{10.1016/j.sbi.2004.05.004}.

\bibitem[Bishop(2006)]{bishop_pattern_2006}
C.~M. Bishop.
\newblock \emph{Pattern recognition and machine learning}.
\newblock Springer, New York, NY, 2006.
\newblock ISBN 0-387-31073-8 978-0-387-31073-2.

\bibitem[Castelo and Roverato(2009)]{castelo_reverse_2009}
R.~Castelo and A.~Roverato.
\newblock Reverse engineering molecular regulatory networks from microarray
  data with qp-graphs.
\newblock \emph{Journal of Computational Biology}, 16\penalty0 (2):\penalty0
  213--227, Feb. 2009.
\newblock ISSN 1066-5277, 1557-8666.
\newblock \doi{10.1089/cmb.2008.08TT}.

\bibitem[Clauset et~al.(2009)Clauset, Shalizi, and
  Newman]{clauset_power-law_2009}
A.~Clauset, C.~R. Shalizi, and M.~E.~J. Newman.
\newblock Power-law distributions in empirical data.
\newblock \emph{SIAM Review}, 51\penalty0 (4):\penalty0 661--703, Nov. 2009.
\newblock ISSN 0036-1445, 1095-7200.
\newblock \doi{10.1137/070710111}.
\newblock arXiv: 0706.1062.

\bibitem[Erdős and Rényi(1959)]{erdos_random_1959}
P.~Erdős and A.~Rényi.
\newblock On random graphs {I}.
\newblock \emph{Publ. Math. Debrecen}, 6:\penalty0 290--297, 1959.

\bibitem[Friedman et~al.(2010{\natexlab{a}})Friedman, Hastie, and
  Tibshirani]{friedman_note_2010}
J.~Friedman, T.~Hastie, and R.~Tibshirani.
\newblock A note on the group lasso and a sparse group lasso.
\newblock \emph{arXiv preprint}, 2010{\natexlab{a}}.
\newblock arXiv:1001.0736.

\bibitem[Friedman et~al.(2010{\natexlab{b}})Friedman, Hastie, and
  Tibshirani]{friedman_regularization_2010}
J.~Friedman, T.~Hastie, and R.~Tibshirani.
\newblock Regularization {Paths} for {Generalized} {Linear} {Models} via
  {Coordinate} {Descent}.
\newblock \emph{Journal of Statistical Software}, 33\penalty0 (1),
  2010{\natexlab{b}}.
\newblock ISSN 1548-7660.
\newblock \doi{10.18637/jss.v033.i01}.

\bibitem[Geman and Geman(1984)]{geman_stochastic_1984}
S.~Geman and D.~Geman.
\newblock Stochastic {Relaxation}, {Gibbs} {Distributions}, and the {Bayesian}
  {Restoration} of {Images}.
\newblock \emph{IEEE Transactions on Pattern Analysis and Machine
  Intelligence}, PAMI-6\penalty0 (6):\penalty0 721--741, Nov. 1984.
\newblock ISSN 0162-8828.
\newblock \doi{10.1109/TPAMI.1984.4767596}.

\bibitem[George and McCulloch(1997)]{george_approaches_1997}
E.~I. George and R.~E. McCulloch.
\newblock Approaches for {Bayesian} variable selection.
\newblock \emph{Statistica Sinica}, 7\penalty0 (2):\penalty0 339--373, Apr.
  1997.
\newblock ISSN 1017-0405.

\bibitem[Geweke(1994)]{geweke_variable_1994}
J.~F. Geweke.
\newblock Variable selection and model comparison in regression.
\newblock Working {Paper} 539, University of Minnesota and Federal Reserve Bank
  of Minneapolis, Minneapolis, MN, Nov. 1994.
\newblock retrieved from https://www.minneapolisfed.org/research/wp/wp539.pdf.

\bibitem[Hager(1989)]{hager_updating_1989}
W.~W. Hager.
\newblock Updating the {Inverse} of a {Matrix}.
\newblock \emph{SIAM Review}, 31\penalty0 (2):\penalty0 221--239, June 1989.
\newblock ISSN 0036-1445, 1095-7200.
\newblock \doi{10.1137/1031049}.

\bibitem[Hastie et~al.(2009)Hastie, Tibshirani, and
  Friedman]{hastie_elements_2009}
T.~Hastie, R.~Tibshirani, and J.~Friedman.
\newblock \emph{The elements of statistical learning: data mining, inference,
  and prediction}.
\newblock Springer {Series} in {Statistics}. Springer, New York, NY, 2nd
  edition, 2009.
\newblock ISBN 978-0-387-84857-0.

\bibitem[Haury et~al.(2012)Haury, Mordelet, Vera-Licona, and
  Vert]{haury_tigress:_2012}
A.-C. Haury, F.~Mordelet, P.~Vera-Licona, and J.-P. Vert.
\newblock {TIGRESS}: trustful inference of gene regulation using stability
  selection.
\newblock \emph{BMC Systems Biology}, 6\penalty0 (1):\penalty0 145, 2012.

\bibitem[Hernández-Lobato(2009)]{hernandez-lobato_prediction_2009}
D.~Hernández-Lobato.
\newblock \emph{Prediction based on averages over automatically induced
  learners: {Ensemble} methods and {Bayesian} techniques}.
\newblock PhD thesis, Universidad Autónoma de Madrid, Madrid, Spain, 2009.
\newblock https://repositorio.uam.es/handle/10486/4161.

\bibitem[Hernández-Lobato et~al.(2013)Hernández-Lobato, Hernández-Lobato,
  and Dupont]{hernandez-lobato_generalized_2013}
D.~Hernández-Lobato, J.~M. Hernández-Lobato, and P.~Dupont.
\newblock Generalized spike-and-slab priors for {Bayesian} group feature
  selection using expectation propagation.
\newblock \emph{Journal of Machine Learning Research}, 14\penalty0
  (1):\penalty0 1891--1945, July 2013.

\bibitem[Hernández-Lobato(2010)]{hernandez-lobato_balancing_2010}
J.~M. Hernández-Lobato.
\newblock \emph{Balancing flexibility and robustness in machine learning:
  {Semi}-parametric methods and sparse linear models}.
\newblock {PhD} thesis, Universidad Autónoma de Madrid, Madrid, Spain, 2010.
\newblock https://repositorio.uam.es/handle/10486/6080.

\bibitem[Hernández-Lobato et~al.(2015)Hernández-Lobato, Hernández-Lobato,
  and Suárez]{hernandez-lobato_expectation_2015}
J.~M. Hernández-Lobato, D.~Hernández-Lobato, and A.~Suárez.
\newblock Expectation propagation in linear regression models with
  spike-and-slab priors.
\newblock \emph{Machine Learning}, 99\penalty0 (3):\penalty0 437--487, June
  2015.
\newblock ISSN 0885-6125, 1573-0565.
\newblock \doi{10.1007/s10994-014-5475-7}.

\bibitem[Karlebach and Shamir(2008)]{karlebach_modelling_2008}
G.~Karlebach and R.~Shamir.
\newblock Modelling and analysis of gene regulatory networks.
\newblock \emph{Nature Reviews Molecular Cell Biology}, 9\penalty0
  (10):\penalty0 770--780, Oct. 2008.
\newblock ISSN 1471-0072, 1471-0080.
\newblock \doi{10.1038/nrm2503}.

\bibitem[Kuo and Mallick(1998)]{kuo_variable_1998}
L.~Kuo and B.~Mallick.
\newblock Variable selection for regression models.
\newblock \emph{Sankhyā: The Indian Journal of Statistics, Series B
  (1960-2002)}, 60\penalty0 (1):\penalty0 65--81, Apr. 1998.
\newblock ISSN 0581-5738.

\bibitem[Kyung et~al.(2010)Kyung, Gill, Ghosh, and
  Casella]{kyung_penalized_2010}
M.~Kyung, J.~Gill, M.~Ghosh, and G.~Casella.
\newblock Penalized regression, standard errors, and {Bayesian} lassos.
\newblock \emph{Bayesian Analysis}, 5\penalty0 (2):\penalty0 369--411, June
  2010.
\newblock ISSN 1936-0975.
\newblock \doi{10.1214/10-BA607}.

\bibitem[Lauritzen(1996)]{lauritzen_graphical_1996}
S.~L. Lauritzen.
\newblock \emph{Graphical {Models}}.
\newblock Clarendon Press, May 1996.
\newblock ISBN 978-0-19-159122-8.
\newblock Google-Books-ID: mGQWkx4guhAC.

\bibitem[Li and Zhang(2017)]{li_bayesian_2017}
F.-q. Li and X.-s. Zhang.
\newblock Bayesian lasso with neighborhood regression method for {Gaussian}
  graphical model.
\newblock \emph{Acta Mathematicae Applicatae Sinica, English Series},
  33\penalty0 (2):\penalty0 485--496, Apr. 2017.
\newblock ISSN 0168-9673, 1618-3932.
\newblock \doi{10.1007/s10255-017-0676-z}.

\bibitem[Lin et~al.(2017)Lin, Wang, Yang, and Zhao]{lin_joint_2017}
Z.~Lin, T.~Wang, C.~Yang, and H.~Zhao.
\newblock On joint estimation of {Gaussian} graphical models for spatial and
  temporal data.
\newblock \emph{Biometrics}, 73\penalty0 (3):\penalty0 769--779, Sept. 2017.
\newblock ISSN 0006-341X.
\newblock \doi{10.1111/biom.12650}.

\bibitem[Liquet et~al.(2017)Liquet, Mengersen, Pettitt, and
  Sutton]{liquet_bayesian_2017}
B.~Liquet, K.~Mengersen, A.~N. Pettitt, and M.~Sutton.
\newblock Bayesian variable selection regression of multivariate responses for
  group data.
\newblock \emph{Bayesian Analysis}, 12\penalty0 (4):\penalty0 1039--1067, Dec.
  2017.
\newblock ISSN 1936-0975, 1931-6690.
\newblock \doi{10.1214/17-BA1081}.

\bibitem[Marbach et~al.(2012)Marbach, Costello, Küffner, Vega, Prill, Camacho,
  Allison, {The DREAM5 Consortium}, Kellis, Collins, and
  Stolovitzky]{marbach_wisdom_2012}
D.~Marbach, J.~C. Costello, R.~Küffner, N.~M. Vega, R.~J. Prill, D.~M.
  Camacho, K.~R. Allison, {The DREAM5 Consortium}, M.~Kellis, J.~J. Collins,
  and G.~Stolovitzky.
\newblock Wisdom of crowds for robust gene network inference.
\newblock \emph{Nature Methods}, 9\penalty0 (8):\penalty0 796--804, Aug. 2012.
\newblock ISSN 1548-7091.
\newblock \doi{10.1038/nmeth.2016}.

\bibitem[Markowetz and Spang(2007)]{markowetz_inferring_2007}
F.~Markowetz and R.~Spang.
\newblock Inferring cellular networks – a review.
\newblock \emph{BMC Bioinformatics}, 8\penalty0 (Suppl 6):\penalty0 S5, 2007.
\newblock ISSN 14712105.
\newblock \doi{10.1186/1471-2105-8-S6-S5}.

\bibitem[Meinshausen and Bühlmann(2006)]{meinshausen_high-dimensional_2006}
N.~Meinshausen and P.~Bühlmann.
\newblock High-dimensional graphs and variable selection with the lasso.
\newblock \emph{The Annals of Statistics}, 34\penalty0 (3):\penalty0
  1436--1462, June 2006.
\newblock ISSN 0090-5364.
\newblock \doi{10.1214/009053606000000281}.

\bibitem[Meinshausen and Bühlmann(2010)]{meinshausen_stability_2010}
N.~Meinshausen and P.~Bühlmann.
\newblock Stability selection.
\newblock \emph{Journal of the Royal Statistical Society: Series B (Statistical
  Methodology)}, 72\penalty0 (4):\penalty0 417--473, Aug. 2010.

\bibitem[Minka(2001{\natexlab{a}})]{minka_expectation_2001}
T.~P. Minka.
\newblock Expectation propagation for approximate {Bayesian} inference.
\newblock In \emph{Proceedings of the {Seventeenth} conference on {Uncertainty}
  in artificial intelligence}, pages 362--369. Morgan Kaufmann Publishers Inc.,
  2001{\natexlab{a}}.

\bibitem[Minka(2001{\natexlab{b}})]{minka_family_2001}
T.~P. Minka.
\newblock \emph{A family of algorithms for approximate {Bayesian} inference}.
\newblock PhD thesis, Massachusetts Institute of Technology, Cambridge, MA,
  2001{\natexlab{b}}.
\newblock https://dspace.mit.edu/handle/1721.1/86583.

\bibitem[Minka and Lafferty(2002)]{minka_expectation-propagation_2002}
T.~P. Minka and J.~Lafferty.
\newblock Expectation-propagation for the generative aspect model.
\newblock In \emph{Proceedings of the {Eighteenth} {Conference} on
  {Uncertainty} in {Artificial} {Intelligence}}, pages 352--359, San Francisco,
  CA, 2002. Morgan Kaufmann Publishers Inc.
\newblock ISBN 978-1-55860-897-9.

\bibitem[Mitchell and Beauchamp(1988)]{mitchell_bayesian_1988}
T.~J. Mitchell and J.~J. Beauchamp.
\newblock Bayesian variable selection in linear regression.
\newblock \emph{Journal of the American Statistical Association}, 83\penalty0
  (404):\penalty0 1023--1032, Dec. 1988.
\newblock ISSN 0162-1459.
\newblock \doi{10.2307/2290129}.

\bibitem[Murphy(2012)]{murphy_machine_2012}
K.~P. Murphy.
\newblock \emph{Machine learning: a probabilistic perspective}.
\newblock Adaptive computation and machine learning series. MIT Press,
  Cambridge, MA, 2012.
\newblock ISBN 978-0-262-01802-9.

\bibitem[Park and Casella(2008)]{park_bayesian_2008}
T.~Park and G.~Casella.
\newblock The {Bayesian} lasso.
\newblock \emph{Journal of the American Statistical Association}, 103\penalty0
  (482):\penalty0 681--686, June 2008.
\newblock ISSN 0162-1459, 1537-274X.
\newblock \doi{10.1198/016214508000000337}.

\bibitem[{R Core Team}(2017)]{r_core_team_r:_2017}
{R Core Team}.
\newblock R: {A} {Language} and {Environment} for {Statistical} {Computing},
  2017.

\bibitem[Seeger(2008)]{seeger_bayesian_2008}
M.~W. Seeger.
\newblock Bayesian inference and optimal design for the sparse linear model.
\newblock \emph{Journal of Machine Learning Research}, 9\penalty0
  (Apr):\penalty0 759--813, Apr. 2008.

\bibitem[Simon et~al.(2013)Simon, Friedman, Hastie, and
  Tibshirani]{simon_sparse-group_2013}
N.~Simon, J.~Friedman, T.~Hastie, and R.~Tibshirani.
\newblock A sparse-group lasso.
\newblock \emph{Journal of Computational and Graphical Statistics}, 22\penalty0
  (2):\penalty0 231--245, Apr. 2013.

\bibitem[{The Gene Ontology Consortium}(2017)]{thegeneontologyconsortium_expansion_2017}
{The Gene Ontology Consortium}.
\newblock Expansion of the {Gene} {Ontology} knowledgebase and resources.
\newblock \emph{Nucleic Acids Research}, 45\penalty0 (D1):\penalty0 D331--D338,
  Jan. 2017.
\newblock ISSN 1362-4962.
\newblock \doi{10.1093/nar/gkw1108}.

\bibitem[Tibshirani(1996)]{tibshirani_regression_1996}
R.~Tibshirani.
\newblock Regression shrinkage and selection via the lasso.
\newblock \emph{Journal of the Royal Statistical Society. Series B
  (Methodological)}, 58\penalty0 (1):\penalty0 267--288, 1996.
\newblock ISSN 0035-9246.

\bibitem[Xu and Ghosh(2015)]{xu_bayesian_2015}
X.~Xu and M.~Ghosh.
\newblock Bayesian variable selection and estimation for group lasso.
\newblock \emph{Bayesian Analysis}, 10\penalty0 (4):\penalty0 909--936, Dec.
  2015.
\newblock ISSN 1936-0975.
\newblock \doi{10.1214/14-BA929}.

\bibitem[Yang and Zou(2015)]{yang_fast_2015}
Y.~Yang and H.~Zou.
\newblock A {Fast} {Unified} {Algorithm} for {Solving} {Group}-lasso {Penalize}
  {Learning} {Problems}.
\newblock \emph{Statistics and Computing}, 25\penalty0 (6):\penalty0
  1129--1141, Nov. 2015.
\newblock ISSN 0960-3174.
\newblock \doi{10.1007/s11222-014-9498-5}.

\bibitem[Yuan and Lin(2006)]{yuan_model_2006}
M.~Yuan and Y.~Lin.
\newblock Model selection and estimation in regression with grouped variables.
\newblock \emph{Journal of the Royal Statistical Society: Series B (Statistical
  Methodology)}, 68\penalty0 (1):\penalty0 49--67, 2006.

\bibitem[Zadeh(1965)]{zadeh_fuzzy_1965}
L.~A. Zadeh.
\newblock Fuzzy sets.
\newblock \emph{Information and Control}, 8\penalty0 (3):\penalty0 338--353,
  June 1965.
\newblock ISSN 0019-9958.
\newblock \doi{10.1016/S0019-9958(65)90241-X}.

\end{thebibliography}

\end{document}